\documentclass[journal]{IEEEtran}
\usepackage{amssymb}
\usepackage{array}
\usepackage{graphicx}
\usepackage{cases}
\usepackage{color}

\usepackage[ruled]{algorithm2e}

\usepackage{caption,setspace}
\usepackage{caption}
\usepackage{multicol}
\usepackage{multirow}
\usepackage{marvosym}
\usepackage{url}
\usepackage{subfig}
\usepackage{epstopdf}
\usepackage{booktabs}
\usepackage{bm}
\usepackage{threeparttable}
\usepackage{cite}

\ifCLASSINFOpdf

\else

\fi

\begin{document}
\title{Multi-task multi-constraint differential evolution with elite-guided knowledge transfer for coal mine integrated energy system dispatching}
\author{Canyun Dai, Xiaoyan Sun \Letter, Hejuan Hu, Wei Song, Yong Zhang, Dunwei Gong
\thanks{ Canyun Dai is with School of Information and Control Engineering, China University of Mining and Technology, Xuzhou, China.
(E-mail:  TB21060011B4@cumt.edu.cn)}
\thanks{\Letter  Xiaoyan Sun  is with School of Information and Control Engineering, China University of Mining and Technology, Xuzhou, China. (E-mail:  xysun78@126.com)}
\thanks{Hejuan Hu is with School of Information and Control Engineering, China University of Mining and Technology, Xuzhou, China. (E-mail: 1101481742@qq.com)}
\thanks{Wei Song is with School of Artificial Intelligence and Computer Science, Jiangnan University, Wuxi, China. (E-mail: songwei@jiangnan.edu.cn)}
\thanks{Yong Zhang is with School of Information and Control Engineering, China University of Mining and Technology, Xuzhou, China. (yongzh401@126.com)}
\thanks{Dunwei Gong  is with School of Information Science and Technology, Qingdao University of Science and Technology, Qingdao, China. (dwgong@vip.163.com)}

\thanks{Manuscript received **; revised **.}}

\markboth{Journal of \LaTeX\ Class Files}%
{Shell \MakeLowercase{\textit{et al.}}: Bare Demo of IEEEtran.cls for IEEE Journals}

\maketitle

\begin{abstract}
The dispatch optimization of coal mine integrated energy system is challenging due to high dimensionality, strong coupling constraints, and multi-objective. Existing constrained multi-objective evolutionary algorithms struggle with locating multiple small and irregular feasible regions, making them inapplicable to this problem. To address this issue, we here develop a multi-task evolutionary algorithm framework that incorporates the dispatch-correlated domain knowledge to effectively deal with strong constraints and multi-objective optimization. Possible evolutionary multi-task construction strategy based on complex constraint relationship analysis and handling, i.e., constraint-coupled spatial decomposition, constraint strength classification and constraint handling technique, is first explored. Within the multi-task evolutionary optimization framework, two strategies, i.e., an elite-guided knowledge transfer by designing a special crowding distance mechanism to select dominant individuals from each task, and an adaptive neighborhood technology-based mutation to effectively balance the diversity and convergence of each optimized task for the differential evolution algorithm, are further developed. The performance of the proposed algorithm in feasibility, convergence, and diversity is demonstrated in a case study of a coal mine integrated energy system by comparing with CPLEX solver and seven state-of-the-art constrained multi-objective evolutionary algorithms.

\end{abstract}


\IEEEpeerreviewmaketitle

\section{Introduction}\label{Int}
\IEEEPARstart{W}{ith} the rapid development in economic and social spheres, issues like fossil energy crisis, ecological deterioration, and global warming have gained increasing prominence. Given this context, there is an urgent need to revolutionize the production and consumption of energy. Integrated energy systems (IES) have emerged as a novel, sustainable, and eco-friendly approach to energy supply, garnering significant interest and successful application in various light industry fields such as communities, islands, and ships \cite{wu2016integrated,boza2021artificial}. However, in light of the proposed two-carbon target, scholars have shifted their focus towards integrated energy systems for high-energy-consuming and high-emission industries. Among these sectors, the coal mining industry has captured substantial attention from both academia and industry. While some studies have initiated relevant research on coal mine integrated energy systems (CMIES), these achievements are still in their nascent stages \cite{hu2022enhanced}.

\par
\begin{figure}[!htb]
\centering	
\includegraphics[width=8.5cm, height=5.5cm]{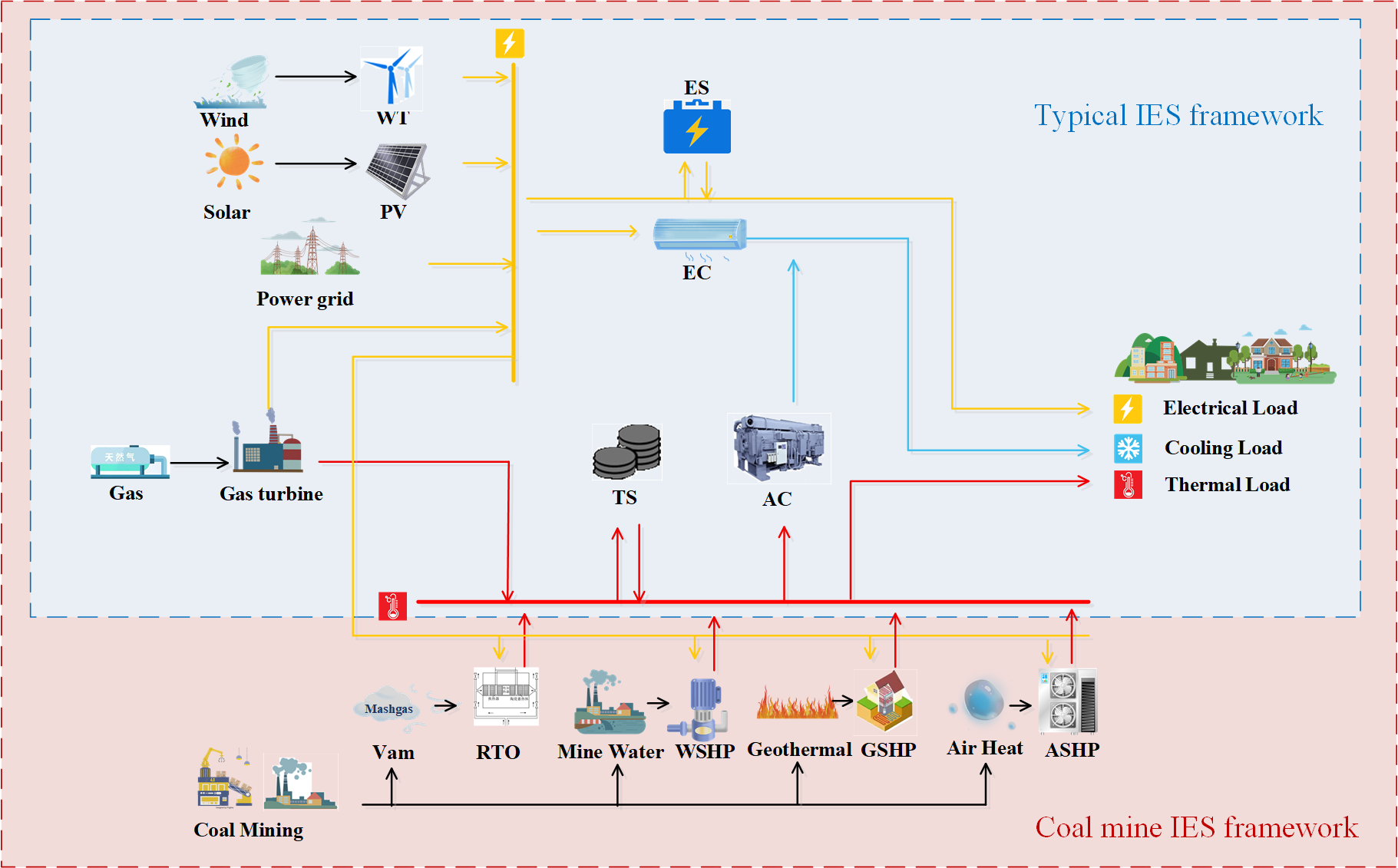}	
\caption{Comparison of typical IES and CMIES frameworks}
\label{CMIES}
\end{figure}

 In contrast to those conventional IES, the coal mine integrated energy system exhibits distinctive characteristics. First, the coal mining process generates a significant amount of associated energy in the form of rich heat, including air heat, ventilation air methane, mine water, and geothermal energy. Second, through the use of specific equipment such as air source heat pumps, ventilation air methane oxidation devices, water source heat pumps, and ground source heat pumps, the thermal energy from these sources can be harnessed to meet the production and living requirements in mining areas. Figure \ref{CMIES} illustrates a comparison between the framework of a typical IES (depicted within the blue dashed box) and a CMIES (depicted within the red dotted line box). As shown in the figure, the CMIES encompasses larger number of sources and conversion devices, leading to a more intricate coupling relationship. This presents a significant challenge for the energy management and optimization of the CMIES.

\par Dispatch optimization has become a focal research in integrated energy systems, holding great significance for system safety, economy, and environmental protection. Studies have developed various dispatch models under different scenarios, primarily categorized as single-objective dispatch models \cite{zhou2019operation,wang2019operation,zhang2021day} and multi-objective ones \cite{song2021economic,zhang2021multi,wang2018optimal}. Single-objective dispatch models typically use commercial optimization solvers such as GUROBI or CPLEX. However, when dealing with multiple optimization objectives, commercial solvers often fail to provide diverse dispatch solutions in a single run. In recent years, population-based evolutionary algorithms (EAs), such as the non-dominated sorting genetic algorithm II (NSGA-II), have been tried to solve multi-objective dispatch problems due to their outstanding performance in obtaining a set of non-dominated solutions with guaranteed convergence, diversity and distribution. Wu et al. \cite{wu2023multi} applied the NSGA-II to solve a multi-objective dispatch model for a non-linear and non-convex park-level integrated energy system. Li et al. \cite{li2018multi} designed a preference-inspired coevolutionary algorithm for an island-level integrated energy system. Wu et al. \cite{wu2021multitasking} proposed an improved multi-task multi-factor evolutionary algorithm for solving multi-objective dispatch problems in different integrated energy systems. Various approaches have further been developed to handle constraints in multi-objective dispatch optimization. Dong et al. \cite{dong2023hybrid}  developed a displacement-based penalty function combined with a state transition algorithm to avoid local optima. Wu et al. \cite{wu2021multi} proposed an improved constraint dominance principle combined with NSGA-II to rapidly reduce constraint violations and effectively identify the feasible region. Wang et al. \cite{wang2020economic} proposed a dynamic epsilon constraint handling approach combined with NSGA-II to effectively locate the feasible region. In summary, EAs have demonstrated attractive performance in solving the aforementioned multi-objective dispatch problems of various integrated energy systems.

\par Due to the advantage of EAs in solving the dispatch optimization problem of integrated energy systems, some scholars have attempted to employ EAs to address the dispatch optimization problem of the coal mine integrated energy system. For instance, Hu et al. \cite{hu2022enhanced} proposed an enhanced NSGA-II algorithm based on timing relationships to efficiently obtain a set of solutions for multi-objective dispatch. This approach reduced the dimensionality of the problem and simplified constraint complexity by dividing the dispatch period. However, this division is highly subjective. Wang et al. \cite{wang2022unified} developed an autonomous intelligent optimization strategy based on support vector machines and designed three strategies to repair infeasible solutions, thereby improving the convergence of the population under strong constraints. Nevertheless, the proposed method entailed high computational complexity. In our previous work \cite{dai2023constraint}, we developed an evolutionary multi-task (EMT)\cite{gupta2015multifactorial} based method to effectively solve the dispatch problem of the coal mine integrated energy system with low-dimensional multi-objective by designing an auxiliary task together with the dispatch one.  Even EAs-based methods have been applied to optimize the dispatch of coal mine integrated energy system, competitive solutions are still very hard to be obtained, especially when the dispatch scenario is complex.

\par Taking inspiration from the successful application of our EMT-based method \cite{dai2023constraint} in solving low-dimensional cases, we here further design a powerful EMT to solve the high-dimensional multi-objective dispatch of the coal mine integrated energy system. To this end, two issues must be focused, one is the task construction to effectively deal with a large number of strong constraints associated with multi-energy coupling, and the other is efficient information sharing strategy for effectively optimizing the multiple tasks in high-dimensional space.

\par Accordingly, the following three contents will be addressed when exploiting the EMT-based method to solve the high-dimensional dispatch optimization of the coal mine integrated energy system. 1) Constructing a domain-adaptive multi-task based dispatch for the coal mine integrated energy system by deeply analyzing the complex constraints relationships under the complex energy coupling knowledge. 2) Designing a knowledge transfer strategy to enhance problem-solving efficiency and minimize transfer time consumption. 3) Improving EA operators to enhance the evolving performance for high dimensional optimization.

\par The main contributions of our algorithm are as follows:

\par
\begin{itemize}
\item \emph{Developed a domain-adaptive multi-task evolutionary dispatch framework by incorporating the constraint knowledge for the coal mine integrated energy system.} Under this framework, three multi-task construction modes based on complex constraint relationship analysis are demonstrated. It includes constraint-coupled variable space decomposition, constraint strength categorization, and constraint handling techniques.
\item \emph{Designed an elite-guided knowledge transfer strategy based on special crowding distance (EKT-SCD).} For each task, the individuals within the same pareto front are ranked using a special crowding distance and only the top 20\% of elite individuals from each pareto front are selected for knowledge transfer. This strategy balances diversity in both objective and decision spaces while reducing the cost of knowledge transfer.
 \item \emph{Proposed a multi-task multi-constraint differential evolution algorithm with elite-guided knowledge transfer and adaptive neighborhood mutation (MMDE-EKT-ANM).} The mutation mechanism uses an angle-based neighborhood technique in the DE/rand/1 strategy to enhance the ability of differential evolution to escape locally feasible regions in high-dimensional space with strong constraints.
\end{itemize}


\par The rest of this paper is arranged as follows. Section \ref{sec2} introduces the multi-objective dispatch optimization model of the coal mine integrated energy system. Section \ref{sec3} develops a multi-task multi-constraint algorithm framework for the dispatch problem. The designed multi-task multi-constrain differential evolution algorithm with elite-guided knowledge transfer and adaptive neighborhood mutation is stated in Section \ref{sec4}. Section \ref{sec5} carries out the experimental results and analysis. The conclusions and future work are outlined in Section \ref{sec6}.


\section{Multi-objective dispatch optimization model of the coal mine integrated energy system}\label{sec2}

\subsection{Description of the coal mine integrated energy system}\label{sec2.1}
\par Figure \ref{CMIES} illustrates the structure of the coal mine integrated energy system with renewable and associated energy sources \cite{hu2022enhanced}. It consists of power grid, wind turbine (WT), photovoltaic (PV), gas turbine (GT), ventilation air methane oxidation devices (RTO), water source heat pump (WSHP), ground source heat pump (GSHP), air source heat pump (ASHP), electrical chiller (EC), absorption chiller (AC), electric storage (ES) and thermal storage (TS). The electrical load is supplied by grid, WT, PV, GT, and ES. GT, RTO, WSHP, GSHP, ASHP, and HS provide the system thermal load. The cooling load is fulfilled by EC and AC. From the above perspective of energy supply, it can be seen that electrical and cold supply are similar to typical integrated energy systems, while thermal supply is different from typical integrated energy systems because of the use of associated energy generated in mine production.

\subsection{Optimization objectives}\label{sec2.2}

\subsubsection{Minimum operating cost}\label{sec2.12}
\par Operating cost includes energy purchase costs $C_{{\rm{buy}}}$, and device operation and maintenance costs $C_{{\rm{opma}}}$.

\begin{equation}
   Minf_{1}=\sum\limits_{t = 1}^{\rm{T}}{(C_{{\rm{buy}},t}+C_{{\rm{opma}},t})}
\end{equation}

\begin{equation}
   C_{{\rm{buy}},t}= \alpha_{{\rm{grid}},t} P_{{\rm{grid}},t}+ \alpha_{{\rm{gt}},t} P_{{\rm{gt}},t}
\end{equation}

\begin{equation}
\begin{array}{l}
   C_{{\rm{opma}},t}= \alpha_{{\rm{wt}}} P_{{\rm{wt}},t}+ \alpha_{{\rm{pv}}}  P_{{\rm{pv}},t}+
   \alpha_{{\rm{ec}}} Q_{{\rm{ec}},t}\\  \quad \quad \quad+ \alpha_{{\rm{ac}}}  Q_{{\rm{ac}},t}+\alpha_{{\rm{rto}}}  H_{{\rm{rto}},t}+ \alpha_{{\rm{ashp}}}  H_{{\rm{ashp}},t}\\ \quad \quad \quad +
   \alpha_{{\rm{wshp}}}  H_{{\rm{wshp}},t}+ \alpha_{{\rm{gshp}}}  H_{{\rm{gshp}},t}\\ \quad \quad \quad +
   \alpha_{{\rm{es}}} (x_{{\rm{es\_out}},t} P_{{\rm{es\_out}},t}+x_{{\rm{es\_in}},t} P_{{\rm{es\_in}},t})\\  \quad \quad \quad   + \alpha_{{\rm{ts}}} (x_{{\rm{ts\_out}},t} H_{{\rm{ts\_out}},t}+x_{{\rm{ts\_in}},t} H_{{\rm{ts\_in}},t})
     \end{array}
\end{equation}
where ${\rm{T}}$ stands for dispatch period; $\alpha$ is the cost factor; $P$, $H$, $Q$ are the electrical, thermal, and cooling power output of each device, respectively; $x_{{\rm{es,t}}}$ is the charging and discharging state of the electrical storage at time $t$; $x_{{\rm{ts,t}}}$ is the charging and discharging state of the thermal storage  at time $t$.

\subsubsection{Minimum abandoned energy cost}\label{sec2.22}
\par Abandoned energy cost includes abandoned renewable energy cost and abandoned associated energy cost.

\begin{equation}
\begin{array}{l}
   Minf_{2}= \sum\limits_{t = 1}^T(\beta_{{\rm{wt}}} (P_{{\rm{wt}},t}^{{\rm{max}}}- P_{{\rm{wt}},t})+ \beta_{{\rm{pv}}} (P_{{\rm{pv}},t}^{{\rm{max}}}- P_{{\rm{pv}},t})\\  \quad \quad \quad +\beta_{{\rm{rto}}} (H_{{\rm{rto}},t}^{{\rm{max}}}-H_{{\rm{rto}},t})+ \beta_{{\rm{ashp}}} (H_{{\rm{ashp}},t}^{{\rm{max}}}-H_{{\rm{ashp}},t})\\ \quad \quad \quad
   +\beta_{{\rm{wshp}}} (H_{{\rm{wshp}},t}^{{\rm{max}}}-H_{{\rm{wshp}},t})\\ \quad \quad \quad+ \beta_{{\rm{gshp}}} (H_{{\rm{gshp}},t}^{{\rm{max}}}-H_{{\rm{gshp}},t}))
     \end{array}
\end{equation}
where $\beta$ is the penalty cost factor of abandoned energy; $P^{{\rm{max}}}$, $H^{{\rm{max}}}$ represent the upper limit of the output electrical power and thermal power of each device, respectively.

\subsection{Constraints}\label{sec2.3}

\subsubsection{Electrical balance constraint}\label{sec2.13}
\begin{equation}
\begin{array}{l}
P_{{\rm{grid}},t}+P_{{\rm{gt}},t}+P_{{\rm{wt}},t}+P_{{\rm{pv}},t}
+x_{{\rm{es\_out}},t} P_{{\rm{es\_out}},t}\\=P_{{\rm{load}},t}+P_{{\rm{rto}},t}+P_{{\rm{ashp}},t}+P_{{\rm{wshp}},t}+P_{{\rm{gshp}},t}+\\x_{{\rm{es\_in}},t} P_{{\rm{es\_in}},t}+P_{{\rm{ec}},t}
\end{array}
\end{equation}
where $P_{{\rm{load}},t}$ is the electrical load at time $t$.

\subsubsection{Thermal balance constraint}\label{sec2.23}
\begin{equation}
\begin{array}{l}
H_{{\rm{gt}},t}+H_{{\rm{rto}},t}+H_{{\rm{ashp}},t}+H_{{\rm{wshp}},t}+H_{{\rm{gshp}},t}+\\x_{{\rm{ts\_out}},t} H_{{\rm{ts\_out}},t}
=H_{{\rm{load}},t}+H_{{\rm{ac}},t}+x_{{\rm{ts\_in}},t}  H_{{\rm{ts\_in}},t}
\end{array}
\end{equation}
where $H_{{\rm{load}},t}$ is the thermal load at time $t$.

\subsubsection{Cooling balance constraint}\label{sec2.33}

\begin{equation}\label{07}
   Q_{{\rm{ec}},t}+Q_{{\rm{ac}},t}=Q_{{\rm{load}},t}
\end{equation}
where $Q_{{\rm{load}},t}$ is the cooling load at time $t$.
\subsubsection{Device output limit and energy conversion constraint}\label{sec2.43}

\begin{equation}\label{rep6}
\left\{
\begin{array}{lll}
 0\leq P_{{\rm{wt}},t}\leq P_{{\rm{wt}},t}^{{\rm{max}}}\\
0\leq P_{{\rm{pv}},t}\leq P_{{\rm{pv}},t}^{{\rm{max}}}\\
0\leq P_{{\rm{grid}},t}\leq P_{{\rm{grid}},t}^{{\rm{max}}}\\
0\leq P_{{\rm{gt}},t}\leq P_{{\rm{gt}},t}^{{\rm{max}}}, H_{{\rm{gt}},t}=\eta_{{\rm{gt}}} P_{{\rm{gt}},t}\\
P_{{\rm{rto}},t}^{{\rm{min}}}\leq P_{{\rm{rto}},t}\leq P_{{\rm{rto}},t}^{{\rm{max}}}, H_{{\rm{rto}},t}=\eta_{{\rm{rto}}} P_{{\rm{rto}},t} \\
P_{{\rm{ashp}},t}^{{\rm{min}}}\leq P_{{\rm{ashp}},t}\leq P_{{\rm{ashp}},t}^{{\rm{max}}}, H_{{\rm{ashp}},t}=\eta_{{\rm{ashp}}} P_{{\rm{ashp}},t} \\
P_{{\rm{wshp}},t}^{{\rm{min}}}\leq P_{{\rm{wshp}},t}\leq P_{{\rm{wshp}},t}^{{\rm{max}}}, H_{{\rm{wshp}},t}=\eta_{{\rm{wshp}}} P_{{\rm{wshp}},t} \\
P_{{\rm{gshp}},t}^{{\rm{min}}}\leq P_{{\rm{gshp}},t}\leq P_{{\rm{gshp}},t}^{{\rm{max}}}, H_{{\rm{gshp}},t}=\eta_{{\rm{gshp}}} P_{{\rm{gshp}},t} \\
0\leq P_{{\rm{ec}},t}\leq P_{{\rm{ec}},t}^{{\rm{max}}}, Q_{{\rm{ec}},t}=\eta_{{\rm{ec}}} P_{\rm{ec},t}\\
0\leq P_{{\rm{ac}},t}\leq P_{{\rm{ac}},t}^{{\rm{max}}}, Q_{{\rm{ac}},t}=\eta_{{\rm{ac}}} P_{{\rm{ac}},t}\\
\end{array} \right.
\end{equation}
where $P^{{\rm{max}}}$, $P^{{\rm{min}}}$ are the upper and lower limits of the output of each device respectively;  $\eta$ represents the energy conversion coefficient of each device.
\subsubsection{Gas turbine climbing constraint}\label{sec2.53}

\begin{equation}\label{rep6}
\left\{
\begin{array}{lll}
P_{{\rm{gt}},t}-P_{{\rm{gt}},t-1}\leq R_{{\rm{up}}}\\
P_{{\rm{gt}},t-1}-P_{{\rm{gt}},t}\leq R_{{\rm{down}}}\\
\end{array} \right.
\end{equation}
where $R_{{\rm{up}}}$, $R_{{\rm{down}}}$ indicate the upper and lower limits of gas turbine climbing respectively.
\subsubsection{Thermal energy storage constraint}\label{sec2.63}

\begin{equation}\label{rep6}
\left\{
\begin{array}{lll}
x_{{\rm{ts\_out}},t},x_{{\rm{ts\_in}},t}\in\{0,1\}\\
0\leq x_{{\rm{ts\_out}},t}+x_{{\rm{ts\_in}},t} \leq 1\\
0\leq x_{{\rm{ts\_out}},t} H_{{\rm{ts\_out}},t} \leq H_{{\rm{ts\_out}},t}^{{\rm{max}}}\\
0\leq x_{{\rm{ts\_in}},t} H_{{\rm{ts\_in}},t} \leq H_{{\rm{ts\_in}},t}^{{\rm{{\rm{max}}}}}\\
S_{{\rm{ts}},t}^{{\rm{min}}}\leq S_{{\rm{ts}},t}\leq S_{{\rm{ts}},t}^{{\rm{max}}}\\
\end{array} \right.
\end{equation}
where $x_{{\rm{ts\_out}},t}$=1 indicates that the device is in the exothermic state; $x_{{\rm{ts\_in}},t}$=1 indicates that the device is in the thermal storage state;  $H_{{\rm{ts\_out}},t}^{{\rm{max}}}$, $H_{{\rm{ts\_in}},t}^{{\rm{max}}}$ are the maximum thermal release and storage power of the thermal storage respectively; $S_{{\rm{ts}},t}^{{\rm{min}}}$, $S_{{\rm{ts}},t}^{{\rm{max}}}$ are the lower and upper limits of the thermal storage.

\subsubsection{Electric energy storage constraint}\label{sec2.73}

\begin{equation}\label{rep6}
\left\{
\begin{array}{lll}
x_{{\rm{es\_out}},t},x_{{\rm{es\_in}},t}\in\{0,1\}\\
0\leq x_{{\rm{es\_out}},t}+x_{{\rm{es\_in}},t} \leq 1\\
0\leq x_{{\rm{es\_out}},t} P_{{\rm{es\_out}},t} \leq P_{{\rm{es\_out}},t}^{{\rm{max}}}\\
0\leq x_{{\rm{es\_in}},t} P_{{\rm{es\_in}},t} \leq P_{{\rm{es\_in}},t}^{{\rm{max}}}\\
S_{{\rm{es}},t}^{{\rm{min}}}\leq S_{{\rm{es}},t}\leq S_{{\rm{es}},t}^{{\rm{max}}}\\
\end{array} \right.
\end{equation}
where $x_{{\rm{es\_out}},t}$=1 indicates that the device is in the discharge state; $x_{{\rm{es\_in}},t}$=1 indicates that the device is in the power storage state; $P_{{\rm{es\_out}},t}^{{\rm{max}}}$, $P_{{\rm{es\_in}},t}^{{\rm{max}}}$ are the maximum discharge and storage power of the electrical storage  respectively; $S_{{\rm{es}},t}^{{\rm{min}}}$, $S_{{\rm{es}},t}^{{\rm{max}}}$ are the lower and upper limits of the electrical storage.

\subsection{Model analysis}\label{sec2.4}
\par Based on the aforementioned model, a mathematical analysis can be conducted to explore the underlying challenges in achieving a set of non-dominated solutions with multi-objective evolutionary optimization. To provide a general analysis, the symbol \textbf{m} is used to represent the number of associated energy types, and the symbol \textbf{n} represents the number of devices. The dimension of decision space is ${\rm{D}}=$\textbf{n}$\times {\rm{24}}$, and the number of constraints ${\rm{E}}=2\times(\textbf{n}-2)\times {\rm{24}}$. In addition, as the number of \textbf{m} increases, the constraints also increase by $2\times \textbf{m}\times {\rm{24}}$. For example, if we here consider $\textbf{m}=4$ and $\textbf{n}=18$, resulting in an optimized variable with dimension as ${\rm{D}}=18\times 24=432$ and constraints number as ${\rm{E}}=32\times24=768$. Apparently, such a constraint scale is quite difficult for traditional EAs. In summary, compared to the typical integrated energy system, the coal mine one involves a larger number of associated energy types (\textbf{m}) and devices (\textbf{n}), leading to a significant increase in the scale of the optimization problem and the number of constraints, as well as a stronger coupling among the constraints. These factors collectively contribute to the immense challenge of finding feasible solutions.

\section{Multi-task multi-constraint evolutionary dispatch framework }\label{sec3}
\par
\begin{figure}[!htb]
\centering	
\includegraphics[width=7cm, height=9.5cm]{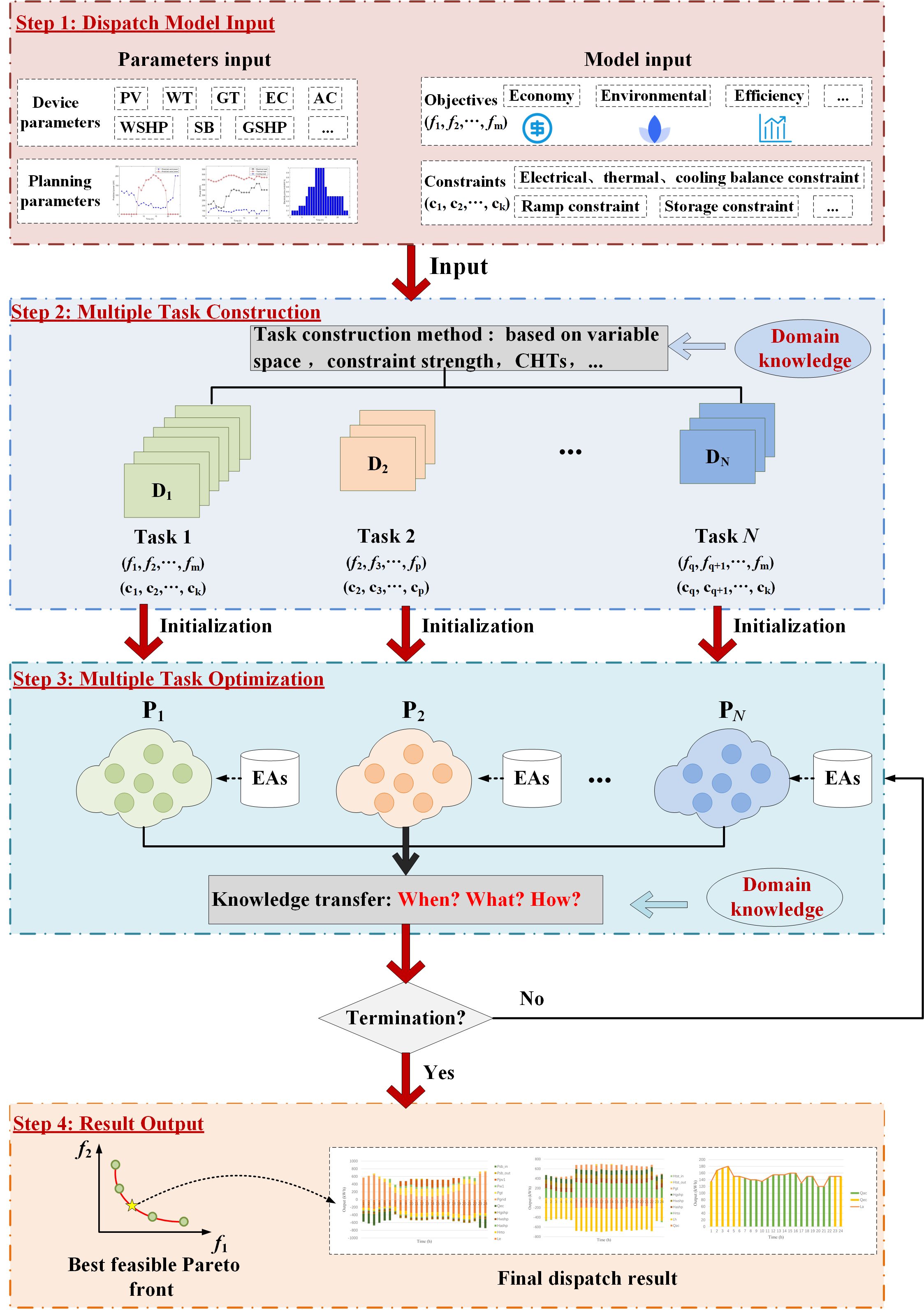}	
\caption{Multi-task multi-constraint evolutionary dispatch algorithm framework}
\label{all}
\end{figure}

 The proposed multi-task based evolutionary framework is shown in Figure \ref{all}. It comprises four main modules: dispatch model input, multi-task construction, multi-task optimization, and result output. Among these, the multi-task construction and multi-task optimization modules are concerned here. For the multi-task construction, strong constraints are essentially managed by analyzing the implicit domain knowledge. Three kinds of constraint relationship analysis will be demonstrated as examples. For the multi-task optimization, an enhanced differential evolution with elite individuals-based knowledge transfer and improved mutation is developed. Clearly, the domain-adaptive task construction is the base of our algorithm, and three alternative methods will be explained in this section. And the specific evolutionary algorithm will be discussed in the following section.

\subsection{Constructed task with constraint-coupled variable space decomposition}\label{sec3.1}
\par
\begin{figure}[!htb]
\centering	
\includegraphics[width=7cm, height=3.6cm]{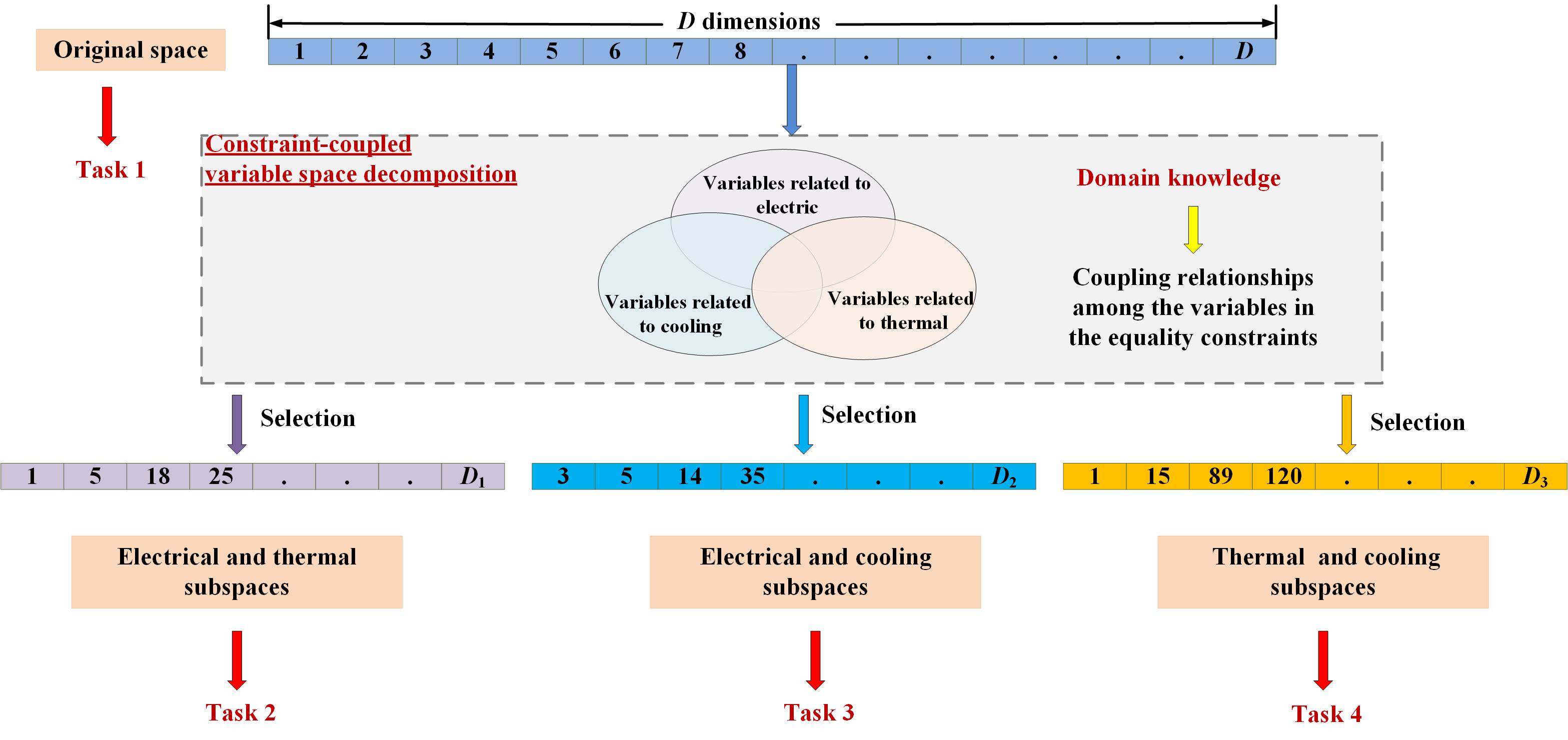}	
\caption{An example of multi-task construction with constraint-coupled variable space decomposition}
\label{space}
\end{figure}

 The constraints of our dispatch are usually strongly coupled due to the multiple energy coupling relationships. However, the coupling relation may often exist among a subset of the energy sources or transformations. According to the physical logic of our problem, we can first recognize the coupling relationships among the optimized variables and then decompose the space into several subspaces. Then multiple tasks with weak and low dimensional constraints can be obtained. Figure \ref{space} provides an example of a multi-task construction with constraint-coupled variable space decomposition. As shown in the figure, according to the coupling relationship between electrical, cooling, and thermal variables under equality constraints, the variable space is decomposed into three low-dimensional subspaces, namely, electrical-thermal subspace, electrical-cooling subspace, and thermal-cooling subspace. Subsequently, these three different subspaces form three different optimization tasks.


\subsection{Constructed task with constraint strength classification }\label{sec3.2}
\par
\begin{figure}[!htb]
\centering	
\includegraphics[width=7cm, height=2cm]{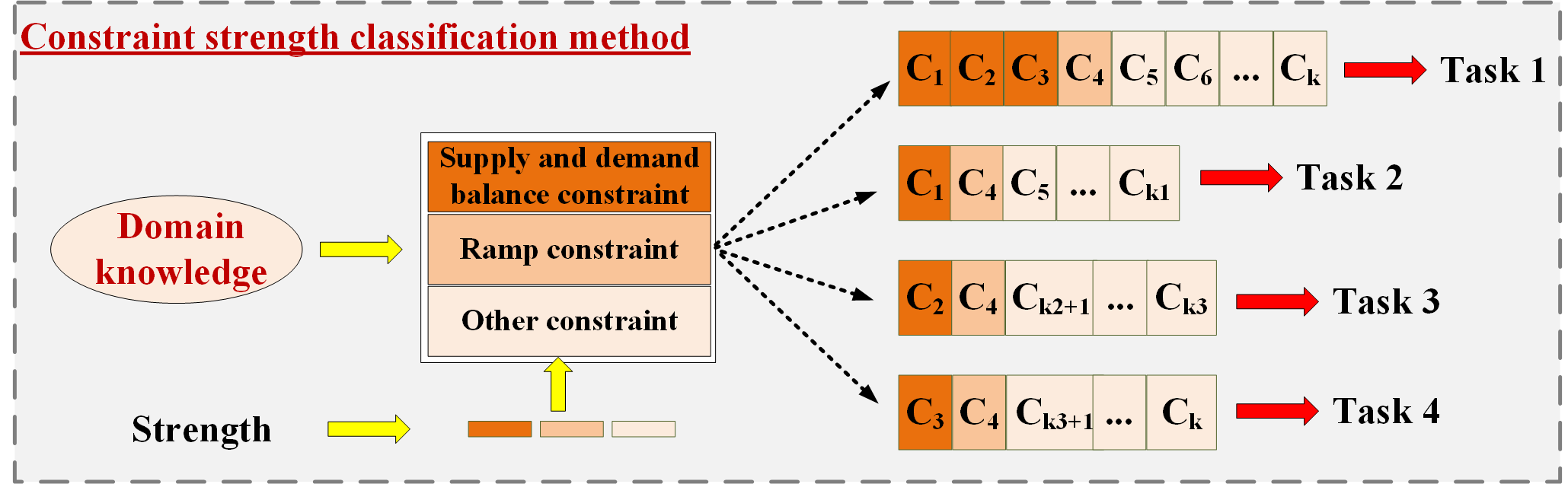}	
\caption{An example of multi-task construction with constraint strength classification}
\label{strength}
\end{figure}

 The strength of each constraint is different in the dispatch of an integrated energy system, for example, the strength of supply and demand balance constraints is larger than that of climbing constraints and other constraints. Therefore, it is logical to classify constraints according to their strength, and then construct multiple optimization tasks with lower intensity. Figure \ref{strength} illustrates an example of constructing multiple tasks based on constraint strength classification. In this example, four tasks are created by combining constraints of different strength levels. Task 1 includes all constraints, Task 2 comprises the electrical balance constraint and partial inequality constraints, Task 3 encompasses the thermal balance constraint and partial inequality constraints, and Task 4 includes the cooling balance constraint and partial inequality constraints. The above task construction method shows that the difficulty of constraint handling can be effectively reduced by assigning the three highest-strength equality constraints to different auxiliary tasks, thus promoting the solution of the problem.

\subsection{Constructed task based on different constraint handling techniques}\label{sec3.3}
\par
\begin{figure}[!htb]
\centering	
\includegraphics[width=7cm, height=2cm]{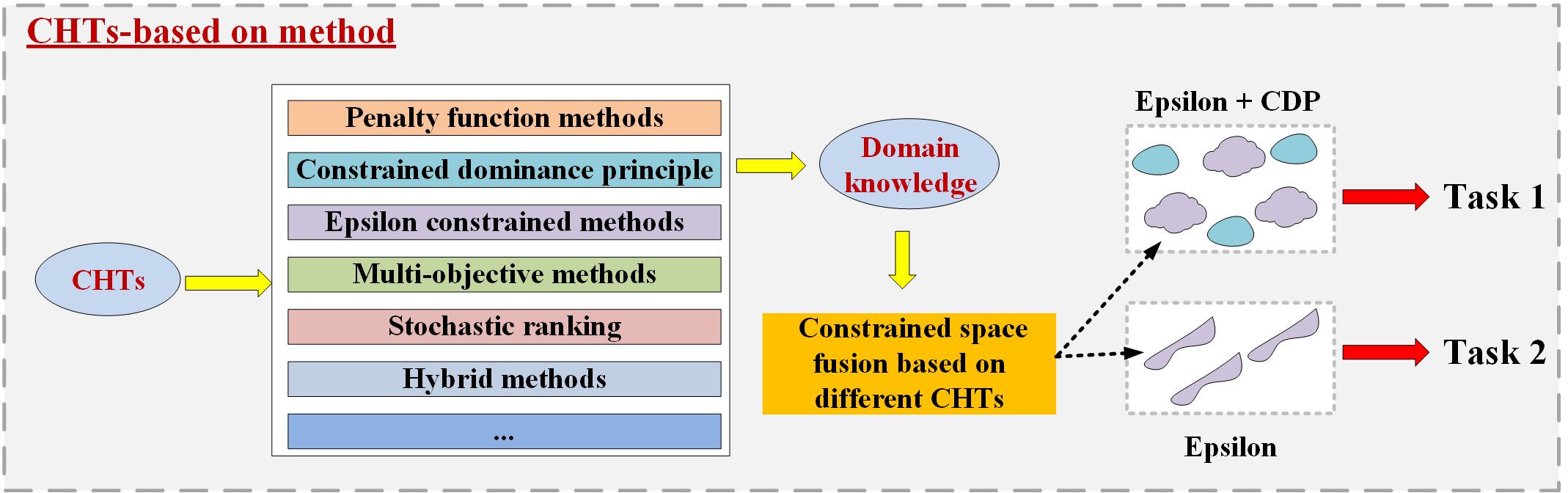}	
\caption{An example of multi-task construction based on CHTs}
\label{CHT}
\end{figure}

 In recent years, various constraint handling techniques (CHTs) have been developed to effectively solve constraints, and the representative ones are the penalty function method \cite{woldesenbet2009constraint}, constraint domination principle (CDP) \cite{deb2000efficient}, epsilon constraint relaxation method \cite{takahama2010efficient} and hybrid method \cite{tasgetiren2010ensemble}. In essence, these methods exhibit different preferences towards constraints, so that different constraint search spaces can be formed. Motivated by this, we propose the construction of multi-tasks by fusing constraint spaces using different CHTs. Figure \ref{CHT} illustrates an example of multi-task construction based on different CHTs, where two optimization tasks are designed according to different CHTs. Task 1 involves the fusion of constraint spaces using a hybrid of CDP and improved epsilon method \cite{qiao2022dynamic}, while Task 2 achieves fusion using the improved epsilon method alone. By employing different CHTs for constraint space fusion, these tasks create different search spaces that help explore multiple different feasible domains.
\par In summary, driven by domain knowledge, we are given three modes to construct multi-tasks based on constraint relationship analysis. These task construction modes may be adapted to different scenario, and the CHTs based method will be adopted in the following from the aspect of performing the multi-task optimization with fewer tasks.
\section{Multi-task multi-constrain differential evolution algorithm with elite-guided knowledge transfer and adaptive neighborhood mutation}\label{sec4}
\par In this section, we propose a multi-task multi-constraint differential evolution algorithm that incorporates elite-guided knowledge transfer and adaptive neighborhood mutation (MMDE-EKT-ANM) to optimize the constructed tasks. The proposed algorithm comprises two key components: an elite-guided knowledge transfer strategy based on special crowding distance (EKT-SCD) and an adaptive neighborhood mutation mechanism (ANM). The main procedure of the proposed MMDE-EKT-ANM algorithm is as demonstrated in Algorithm 1.

\subsection{Main process of MMDE-EKT-ANM}\label{sec4.1}
\par Algorithm 1 presents the pseudocode of MMDE-EKT-ANM. In line 1, initial populations $P_{1}$, $P_{2}$ with $N$ individuals for each task are randomly generated in the search space. The populations are then evaluated, and the iteration counter $G$ is set to 1. Following that, in line 5, the ANM mechanism is executed to generate offspring populations $O_{1}$, $O_{2}$ for each task, with the details explained in Section \ref{sec4.3}. Subsequently, the offspring populations are evaluated for each task. Next, in lines 7-9, the EKT-SCD process is performed, which will be presented in Section \ref{sec4.2}. Finally, an environmental selection operation is applied to the merged population, with Task 1 adopting a hybrid CDP and improved epsilon strategy, while Task 2 uses the improved epsilon strategy. Then, in line 12, the iteration counter $G$ is incremented by 1. If the value of $G$ is less than $G_{{\rm{max}}}$, the process from lines 4-12 is iterated; otherwise, $P_{1}$ is outputted as the final solution set.

\begin{algorithm}
\caption{The pseudo code of  MMDE-EKT-ANM}\label{I} 
\LinesNumbered 
\KwIn{$N$: size of the population, $D$: dimension of the population, $G_{{\rm{max}}}$: maximum number of iterations}
\KwOut{$P_{1}$: the feasible pareto optimal solutions }
$P_{1}$,$P_{2}$ $\leftarrow$ Initialize $N$ individuals for each task\\
Evaluate $P_{1}$,$P_{2}$  \\
$G$ $\leftarrow$ 1\\
\While {$G\leq G_{{\rm{max}}}$}{
$O_{1}$,$O_{2}$ $\leftarrow$ Use the ANM to generate offspring\\
Evaluate $O_{1}$,$O_{2}$  \\
Determine transfer solution $R_{1}$,$R_{2}$ according to the SCD\\
$P_{1}$ $\leftarrow$ $P_{1}\cup O_{1} \cup R_{2}$ (EKT-SCD knowledge transfer)\\
$P_{2}$ $\leftarrow$ $P_{2}\cup O_{2} \cup R_{1}$ (EKT-SCD knowledge transfer)\\
$P_{1}$ $\leftarrow$ CDP+improved epsilon($P_{1}$,$N$)\\
$P_{2}$ $\leftarrow$ improved epsilon($P_{2}$,$N$)\\

$G$ $\leftarrow$ $G$ +1\\
}
\end{algorithm}

\subsection{Elite-guided knowledge transfer based on special crowding distance}\label{sec4.2}
\par
\begin{figure}[!htb]
\centering	
\includegraphics[width=8cm, height=4.5cm]{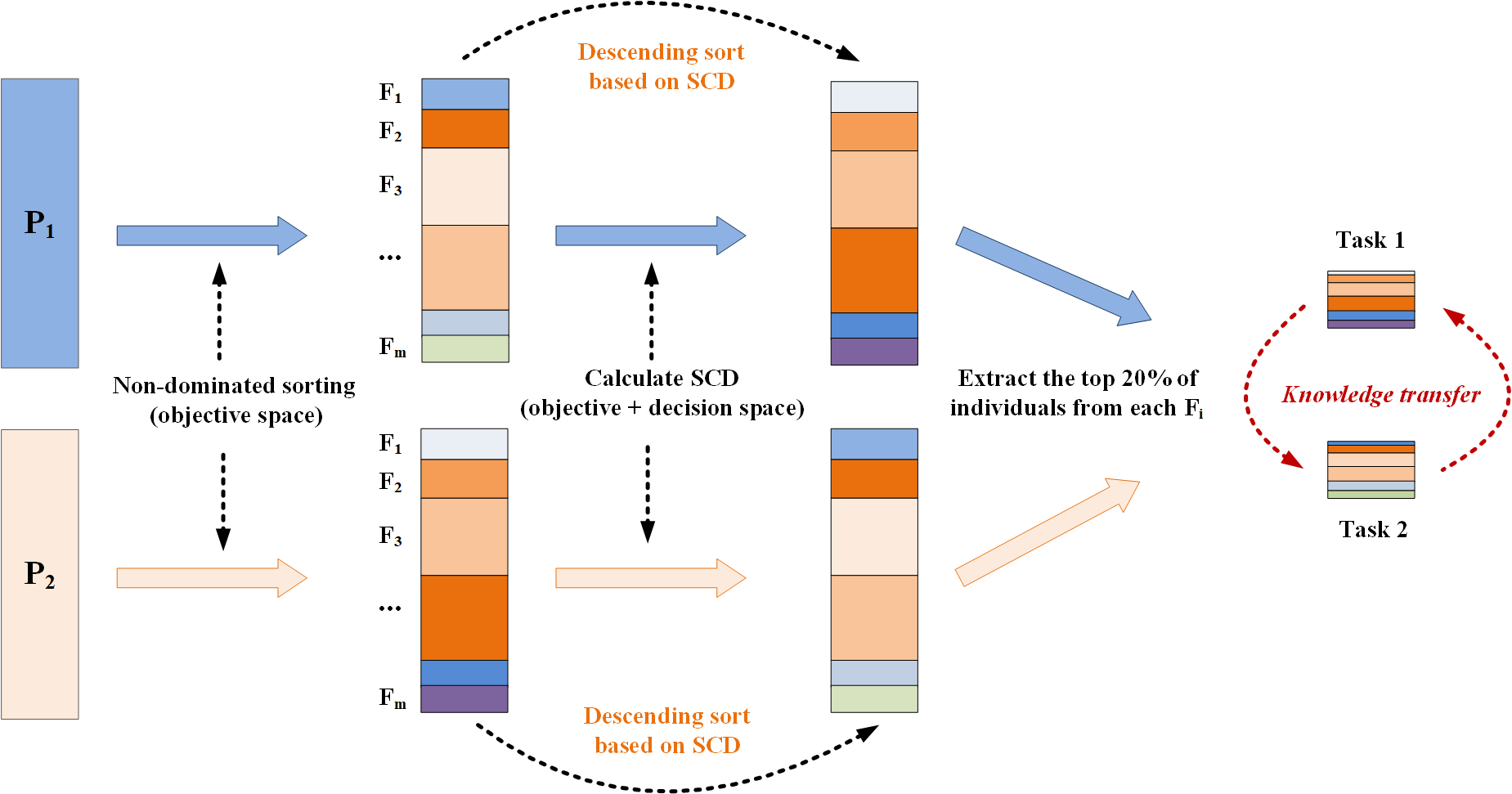}	
\caption{ The process of EKT-SCD strategy}
\label{KT}
\end{figure}

 Negative transfer often occurs in the EMT algorithm, which causes wrong search directions and worse results. To achieve efficient knowledge transfer, we propose an elite-guided knowledge transfer strategy based on special crowding distance presented in \cite{yue2017multiobjective}. First, the individuals in the same front are sorted by calculating the crowding distance of the individuals, and then the top 20\% individuals from each front are selected as elite individuals for knowledge transfer. The process of the proposed strategy is presented in Figure \ref{KT}.

\textbf{\emph{Discuss}}: The EKT-SCD mechanism has the following three advantages. (1) SCD ensures the diversity of transferred knowledge in both the decision space and objective space, thereby assisting the algorithm in obtaining a diverse and well-converged pareto front. (2) The knowledge from the constructed auxiliary task enables the main task to explore diverse search paths, while the knowledge from the main task helps the auxiliary one discover new certain feasible regions. (3) Since knowledge transfer incurs additional time costs, only the top 20\% of the elite individuals are chosen for knowledge transfer during the evolution process, effectively reducing the algorithm's time consumption.

\subsection{Differential evolution based on adaptive neighborhood mutation}\label{sec4.3}
\par Differential evolution (DE) \cite{storn1997differential} is a simple yet powerful EA that has been successfully applied to many optimization problems \cite{sheng2020differential,wang2022surrogate}. In particular, DE has shown promise in dispatch optimization for integrated energy systems. Based on this, we choose the DE as the optimizer to implement the task optimization. However, due to the high-dimensional and multi-constraint nature of the dispatch optimization of the coal mine integrated energy system, the optimization performance of the DE is degraded. To address this, we embed a neighborhood technique \cite{xiao2023locally} with enhanced diversity into DE and propose an adaptive neighborhood mutation mechanism to improve DE performance. Specifically, an angle-based neighborhood strategy \cite{wang2022surrogate} is first used to construct Nr neighborhoods (Nr set to 10) for each individual. Then, within the constructed neighborhoods, the DE/rand/1 strategy \cite{zhou2023clustering} is combined to further enhance the diversity of local search, helping DE to escape local feasible domains. Moreover, the DE/current-to-best/1 strategy \cite{zhou2023clustering} is applied to the entire population to enhance global search convergence and feasibility. To achieve a balance among diversity, convergence, and feasibility, an adaptive mechanism is designed. The ANM mechanism is described as follows:

\begin{equation}\label{rep6}
v_{i}=\left\{
\begin{array}{lll}
x_{i}+F_{i}(x_{{\rm{r1^{'}}}}-x_{{\rm{r2^{'}}}}), if \ {\rm{rand}}_{i}<{\rm{P_{c}}} \\
x_{i}+F_{i}(x_{{\rm{best}}}-x_{i})+F_{i}(x_{{\rm{r1}}}-x_{{\rm{r2}}}), otherwise\\
\end{array} \right.
\end{equation}
where ${\rm{rand}}_{i}$ returns a random number ranging from 0 to 1; ${\rm{r1^{'}}}$, ${\rm{r2^{'}}}$ are two individuals randomly selected from the neighborhood formed by $x_{i}$ and ${\rm{r1^{'}}}\neq {\rm{r2^{'}}}$;  ${\rm{r1}}$, ${\rm{r2}}$ are two individuals randomly selected from the current population and ${\rm{r1}}\neq {\rm{r2}}$; $x_{{\rm{best}}}$ is the best individual; F=\{0.6,0.8,1.0\} is randomly chosen from three different values, which have been widely used in previous literature due to their contributions to diversity and maintaining good search capability \cite{qiao2022dynamic}; $\rm{P_{c}}$  is a probability parameter $\rm{P_{c}}=1-G/G_{\rm{max}}$.

\textbf{\emph{Discuss}}: The inherent advantages of the ANM mechanism are further analyzed in conjunction with the tasks constructed in Section \ref{sec3.3}. (1) By combining the neighborhood and DE/rand/1 strategies, the diversity of local search can be enhanced, prompting Task 1 to locate multiple discrete feasible regions, while Task 2 can thoroughly explore local infeasible regions. (2) The DE/current-to-best/1 strategy enables global search to increase feasibility and convergence speed. In this case, Task 1 can approach excellent individuals, thereby improving population distribution. Task 2 can accelerate the search to discover more promising regions. (3) The adaptive strategy maintains a good balance between diversity, feasibility, and convergence. In the early stages of evolution, most individuals adopt the DE/rand/1 strategy, thereby enhancing the exploration capability and diversity of the population. In the later stages of evolution, as the individuals in the current population have converged near the pareto optimal solutions, they gradually tend to choose the DE/current-to-best/1 strategy to improve search efficiency and accuracy. In summary, the ANM mechanism can locate multiple feasible regions and avoid premature convergence, allowing DE to achieve higher search efficiency.

\section{Application in typical coal mine integrated energy system }\label{sec5}

\subsection{Parameters setting }\label{sec5.1}
\par
\begin{figure}[!htb]
\centering	
\includegraphics[width=7.5cm, height=5.5cm]{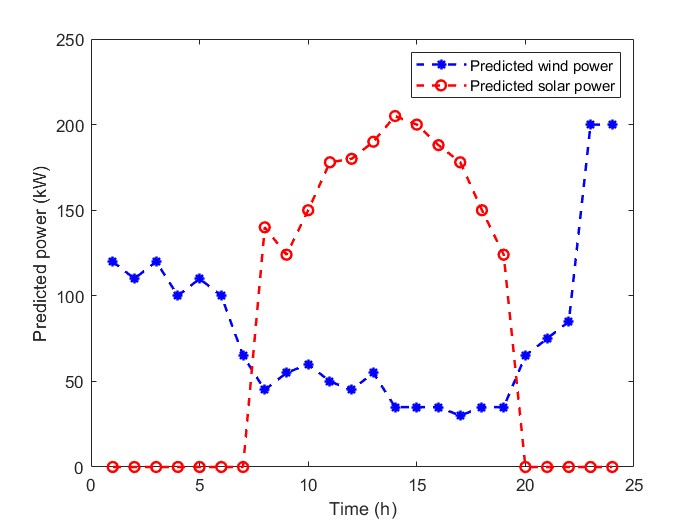}	
\caption{Predicted wind and solar power  curves}
\label{WT}
\end{figure}

\begin{figure}[!htb]
\centering	
\includegraphics[width=7.5cm, height=5.5cm]{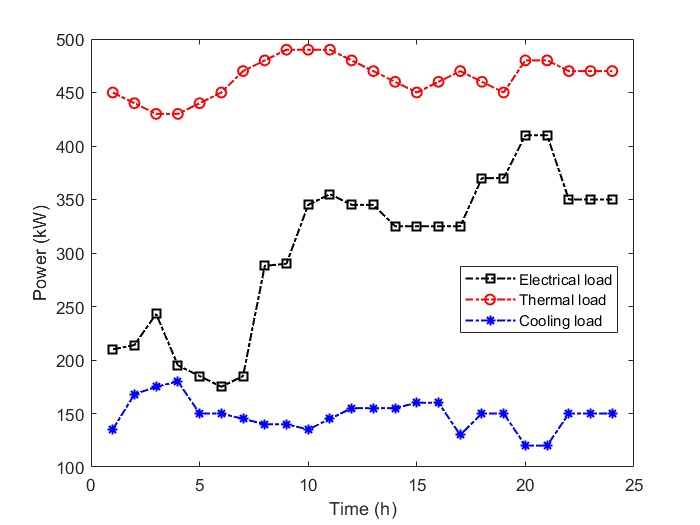}	
\caption{Electrical, thermal and cooling load curves}
\label{load}
\end{figure}

 The feasibility and effectiveness of the proposed algorithm are experimentally demonstrated by applying it to day-ahead dispatch optimization at a specific mine in Shanxi, China. The predicted power of wind ands solar, as well as the predicted loads of electrical, cooling, and thermal are illustrated in Figure \ref{WT} and Figure \ref{load}, respectively. The parameters of various devices are presented in Table \ref{Parameters} and real-time electricity prices can be referenced in \cite{dai2023constraint}. The PlatEMO platform of MATLAB is conducted on a personal computer with an Intel(R) Core i7-11700 2.5 GHz CPU and 16.00 GB RAM \cite{tian2017platemo}.


\begin{table*}[!htb]
 \footnotesize
  \centering
  \caption{Device technical and economic parameters} \label{Parameters}
  \resizebox{\linewidth}{!}{
  \begin{threeparttable}
   \begin{tabular}{llllll}
    \toprule
    \multirow{1}{*} Device & Parameter & Value & Device & Parameter & Value \\[1pt]
   \midrule
    PV  & operational and maintenance cost (rmb/kWh) & $\alpha_{{\rm{pv}}}$=0.32      & RTO  & operational and maintenance cost (rmb/kWh) &   $\alpha_{{\rm{rto}}}$=0.165\\
    & abandoned energy cost (rmb/kWh)   & $\beta_{{\rm{{\rm{pv}}}}}$=0.12   &   & abandoned energy cost (rmb/kWh)   & $\beta_{{\rm{rto}}}$=0.6\\
    & maximum output (kW)     & $P_{{\rm{pv}},t}^{{\rm{max}}}$=predicted value             &  & minimum output (kW)      & $P_{{\rm{rto}},t}^{{\rm{min}}}$=30 \\
    &   &  &  & maximum output (kW)     & $P_{{\rm{rto}},t}^{{\rm{min}}}$ =150 \\
    WT    & operational and maintenance cost (rmb/kWh) & $\alpha_{{\rm{wt}}}$=0.3&  & conversion efficiency of RTO  & $\eta^{{\rm{rto}}}$=3.0\\
    &  abandoned energy cost (rmb/kWh)    & $\beta_{{\rm{wt}}}$=0.1      &    &       \\[1pt]

    &  maximum output (kW)     & $P_{{\rm{wt}},t}^{{\rm{max}}}$=predicted value& WSHP & operational and maintenance cost (rmb/kWh)&$\alpha_{{\rm{wshp}}}$=0.163\\
          & &  &  & abandoned energy cost (rmb/kWh)  &$\beta_{{\rm{wshp}}}$=0.5\\
          ASHP    & operational and maintenance cost (rmb/kWh) & $\alpha_{{\rm{ashp}}}$=0.16&   & minimum output (kW)      & $P_{{\rm{wshp}},t}^{{\rm{min}}}$ =30  \\
          & abandoned energy cost (rmb/kWh)    & $\beta_{{\rm{ashp}}}$=0.52    & & maximum output (kW)& $P_{{\rm{wshp}},t}^{{\rm{max}}}$ =80 \\
          & minimum output (kW)      & $P_{{\rm{ashp}},t}^{{\rm{min}}}$ =30  &  & conversion efficiency of WSHP  & $\eta^{{\rm{wshp}}}$=2.95\\
          &  maximum output (kW)& $P_{{\rm{ashp}},t}^{{\rm{max}}}$ =100  &      &    \\[1pt]

          &conversion efficiency of ASHP  & $\eta^{{\rm{ashp}}}$=2.9& GSHP & operational and maintenance cost (rmb/kWh)&$\alpha_{{\rm{gshp}}}$=0.165\\
          & &  &  & abandoned energy cost (rmb/kWh)  &$\beta_{{\rm{gshp}}}$=0.2\\

          GT& maximum output (kW) & $P_{{\rm{gt}},t}^{{\rm{max}}}$ =350 & & minimum output (kW) & $P_{{\rm{gshp}},t}^{{\rm{min}}}$ =30  \\
          & ramp down (kW) & $R_{{\rm{up}}}$=50 & & maximum output (kW)& $P_{{\rm{gshp}},t}^{{\rm{max}}}$ =80 \\
          & ramp up (kW) &  $R_{{\rm{down}}}$=-50  &  & conversion efficiency of GSHP  & $\eta^{{\rm{gshp}}}$=3.1\\
          & thermoelectric ratio of GT  & $\eta_{{\rm{gt}}}$=0.58  &   & \\[1pt]
          & && Grid  &  maximum output (kW) & $P_{{\rm{grid}},t}^{{\rm{max}}}$ =800 \\
          & & & & &\\

          AC & operational and maintenance cost (rmb/kWh) &$\alpha_{{\rm{ac}}}$=0.3 &EC& operational and maintenance cost (rmb/kWh) & $\alpha_{{\rm{ec}}}$=0.2 \\
           & maximum output (kW)  & $P_{{\rm{ac}},t}^{{\rm{max}}}$ =260 & &   maximum output (kW) & $P_{{\rm{ec}},t}^{{\rm{max}}}$ =280    \\
           &conversion efficiency of AC    & $\eta^{{\rm{ac}}}$=0.7 &   & conversion efficiency of EC  & $\eta^{{\rm{ec}}}$=0.65  \\
           &  &  & &\\

         ES    & operational and maintenance cost (rmb/kWh) & $\alpha_{{\rm{es}}}$=0.2   & TS & operational and maintenance cost (rmb/kWh)&$\alpha_{{\rm{ts}}}$=0.1\\
          &  maximum output (kW)& $P_{{\rm{es}},t}^{{\rm{max}}}$ =30 & & maximum output (kW)& $P_{{\rm{ts}},t}^{{\rm{max}}}$ =30 \\
          & energy storage efficiency of ES  & $\eta^{{\rm{es}}}$=0.98&  & energy storage efficiency of TS  & $\eta^{{\rm{ts}}}$=0.95\\

    \bottomrule
  \end{tabular}
  \footnotesize
  \end{threeparttable}}
\end{table*}

\subsection{Experiments setting }\label{sec5.2}
\par To evaluate the performance of the proposed algorithm, two groups of comparative experiments are conducted. Group 1: the most often used CPLEX solver is compared by transforming the multi-objective dispatch problems into single-objective one through linear weighting. To ensure fairness, the CPLEX solver performs iterations with $N$ sets of different weights varied in the range [0, 1]. Here, $N$ corresponds to the population size of our algorithm. Group 2: seven state-of-the-art constrained multi-objective evolutionary algorithms (CMOEAs), i.e., co-evolutionary CMOEA (CCMO) \cite{tian2020coevolutionary}, dual-population based evolutionary algorithm (c-DPEA) \cite{ming2021dual}, EMT-based constraint multi-objective optimization algorithm (EMCMO) \cite{qiao2022evolutionary}, improved EMCMO algorithm (CMOEMT) \cite{ming2022constrained}, double-balanced EMT algorithm (DBEMTO) \cite{qiao2023self}, dynamic auxiliary task-based on EMT algorithm (MTCMO) \cite{qiao2022dynamic}, and improved MTCMO algorithm (IMTCMO) \cite{qiao2023evolutionary}, are compared. All algorithms are configured with a population size of 300 and a maximum iteration of 5000. To ensure reliability, each algorithm independently runs 20 times.

\par To evaluate the performance of the non-dominated solution sets obtained by each algorithm, the inverted generation distance (IGD) \cite{bosman2003balance} and hypervolume (HV) \cite{liu2020fuzzy} indicators are adopted. The IGD focuses on measuring the proximity between the obtained PF and the optimal PF, reflecting the convergence of the algorithm. A smaller IGD value indicates better convergence. The HV is a comprehensive evaluation indicator that simultaneously assesses the convergence and diversity of the algorithm. A larger HV value indicates better performance in convergence and diversity. It is worth noting that all algorithms share the same reference set for the IGD and HV indicators.

\subsection{Experimental results and analysis}\label{sec5.3}

\subsubsection{Results of Group 1: Compared  with  CPLEX solver}\label{sec5.3.1}
\par Comparing the solution sets derived from our algorithm to those produced by the CPLEX solver using 300 weight settings, we can reasonably determine the feasibility of the  proposed algorithm in addressing the dispatch problem of coal mine integrated energy systems since CPLEX is greatly credible.

\par
\begin{figure}[!htb]
\centering	
\includegraphics[width=8cm, height=6cm]{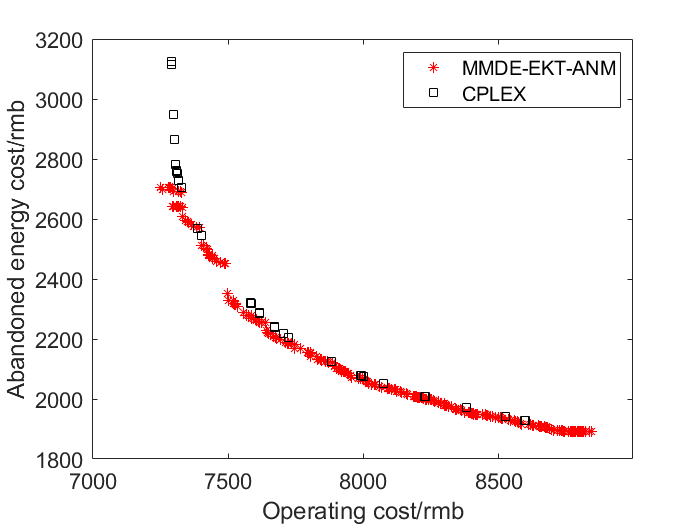}	
\caption{The PF comparison between the proposed algorithm and CPLEX solver}
\label{CPLEX1}
\end{figure}

 Figure \ref{CPLEX1} depicts the distribution of the obtained solutions in the objective space. As can be seen from the figure, although our algorithm is slightly less diverse than the CPLEX solver, it has superior convergence.

\par Numerical analysis is further compared by selecting two endpoints  from Figure \ref{CPLEX1}, and the results are presented in Table \ref{BS1}. For the left endpoint, result of our algorithm clearly outperforms the CPLEX in both reducing the operation cost (save 41.44 rmb) and the abandoned energy cost (save 419.43 rmb). For the right endpoint, the solutions obtained by both methods are non-dominated. Compared to the solution of the CPLEX solver, although the operation cost of our method increases 137.51 rmb, it effectively reduces the cost of abandoned energy cost by 34.76 rmb, thus helping to reduce carbon emissions.  Besides, the runtime of these two algorithms is also compared, the running time of our algorithm is about twice that of the CPLEX solver and it is acceptable. In conclusion, although our algorithm exhibits a slight compromise in terms of runtime and diversity, it outperforms the CPLEX solver in convergence and energy utilization. Therefore, the above conclusions verify the feasibility and effectiveness of the proposed algorithm in solving the coal mine integrated energy system dispatch problem.

\begin{table*}
 \centering
  \caption{Comparison between the proposed algorithm and CPLEX solver at two endpoints and time}
  \label{BS1}
  \begin{tabular}{llllllll}
    \toprule
    \multirow{2}{*}{Approach} & Left endpoint &  & Right endpoint & &  \\[1pt]
    \cline{2-5}
    & CPLEX solver & MMDE-EKT-ANM & CPLEX solver & MMDE-EKT-ANM \\[1pt]
    \midrule
    Operating cost/rmb& 7290.50   &7249.06(-41.44) &8600.69& 8738.20 (+137.51)\\[1pt]
    Abandoned energy cost/rmb &3125.57 & 2706.14(-419.43) &1927.75  &  1892.99(-34.76)\\[1pt]
    Time/s &  507 & & 983 &\\[1pt]
   \bottomrule
  \end{tabular}
\end{table*}

\subsubsection{ Results of Group 2: Compared with the seven state-of-the-art algorithms}\label{sec5.3.2}
\par The values of IGD and HV indicators  are listed in Table \ref{IGD} and Table \ref{HV}, where the ``Mean" displays the average values obtained from 20 runs, providing insights into the convergence and diversity of the non-dominated solution sets. The ``Std" represents the variance  and reflects the stability of the algorithms. The ``Best"  showcases the optimal results achieved after 20 runs, while the ``Worst" presents the corresponding worst values. Optimal results for the IGD and HV indicators are denoted in bold, and suboptimal results are underlined.

\par The following conclusions can be observed from Table \ref{IGD}: (1) The proposed MMDE-EKT-ANM achieves the optimal Mean value as 17.6654 for the IGD indicator, 123.0492 less than that of  the second-ranked CMOEMT algorithm, indicating a prominent convergence. (2) The proposed algorithm exhibits the smallest Std value, amounting to only 28\% of the value obtained by the second-best algorithm, IMTCMO, which represents a more stable optimization performance. (3) MMDE-EKT-ANM also excels in the Best and Worst values for the IGD indicator. It attains 11\% of the second-ranked algorithm IMTCMO for the Best value and 17\% of the second-ranked algorithm MTCMO for the Worst value. In summary, the results strongly indicate that the MMDE-EKT-ANM algorithm significantly surpasses the other compared algorithms in terms of convergence and stability.

\begin{table}
\footnotesize
  \caption{Comparison of IGD values of each algorithm}
  \label{IGD}
  \begin{tabular}{llllllll}
    \toprule
    \multirow{2}{*}{Approach} &  &IGD  & & &  \\[1pt]
    \cline{2-5}
    & Mean & Std & Worst &Best\\[1pt]
    \midrule
CMOEMT\cite{ming2022constrained}&\underline{140.7146} 	&31.5739  	&216.0402   &	99.9542\\[1pt]
DBEMTO\cite{qiao2023self}&289.4024 &	49.0232  	&358.5951  &	201.1625\\[1pt]
EMCMO\cite{qiao2022evolutionary}&268.4530 &	40.8786 & 	351.3277  &	203.1713\\[1pt]
IMTCMO\cite{qiao2023evolutionary}&145.5954 	&\underline{21.3162} 	&\underline{171.3711} &99.5043\\[1pt]
MTCMO\cite{qiao2022dynamic}&186.0881 &	53.4267 & 	271.9225    	&\underline{80.7308}\\[1pt]
CCMO\cite{tian2020coevolutionary}&284.2009 & 	46.8279  & 	389.6458   &	202.4054\\[1pt]
cDPEA\cite{ming2021dual}&269.2086 &	60.6885 & 	402.3848  	&176.0488\\[1pt]
MMDE-EKT-ANM&\textbf{17.6654} 	&\textbf{6.0689 }	&\textbf{28.5303 } 	&\textbf{9.2926}\\[1pt]

   \bottomrule
  \end{tabular}
\end{table}

\par Table \ref{HV} lists the HV results among eight different algorithms. (1) Analysis of Table \ref{HV} reveals that MMDE-EKT-ANM consistently achieves better HV values compared to other compared algorithms. CMOEMT, IMTCMO, and MTCMO secure second-ranked results in Mean, Worst, and Best values, respectively, at only 81\%, 78\%, and 89\% of the proposed algorithm. (2) In terms of the Std value, the proposed algorithm obtains the smallest one, which is only 40\% of the second-ranked IMTCMO. This further confirms the relatively stable performance of MMDE-EKT-ANM. In conclusion, MMDE-EKT-ANM effectively addresses dispatch optimization for the coal mine integrated energy system, outperforming other evolutionary algorithms and delivering superior performance in terms of diversity, convergence, and stability.

\begin{table}
\footnotesize
  \caption{Comparison of HV values of each algorithm}
  \label{HV}
  \begin{tabular}{llllllll}
    \toprule
    \multirow{2}{*}{Approach} &  &HV  & & &  \\[1pt]
    \cline{2-5}
    & Mean & Std & Worst &Best\\[1pt]
    \midrule
CMOEMT\cite{ming2022constrained}&\underline{0.7824} &	0.0418&	0.6825 &	0.8372\\[1pt]
DBEMTO\cite{qiao2023self}&0.6011   &  	0.0561    & 	0.5262     &	0.7087\\[1pt]
EMCMO\cite{qiao2022evolutionary}&0.6200  &    	0.0458    &  	0.5295  &    	0.6964\\[1pt]
IMTCMO\cite{qiao2023evolutionary}&0.7735  &	\underline{0.0279} &\underline{0.7382 }&	0.8376\\[1pt]
MTCMO\cite{qiao2022dynamic}&0.7352     &	0.0647    & 	0.6437  &   	\underline{0.8739}\\[1pt]
CCMO\cite{tian2020coevolutionary}&0.6042    & 	0.0514   &  	0.5008   &  	0.7052\\[1pt]
cDPEA\cite{ming2021dual}&0.6490    &	0.0626    &	0.4986    &	0.7463\\[1pt]
MMDE-EKT-ANM&\textbf{0.9609}  &	\textbf{0.0113} &   	\textbf{0.9445} &   	\textbf{0.9781}\\[1pt]
   \bottomrule
  \end{tabular}
\end{table}

\par
\begin{figure}[!htb]
\centering	
\includegraphics[width=8cm, height=6cm]{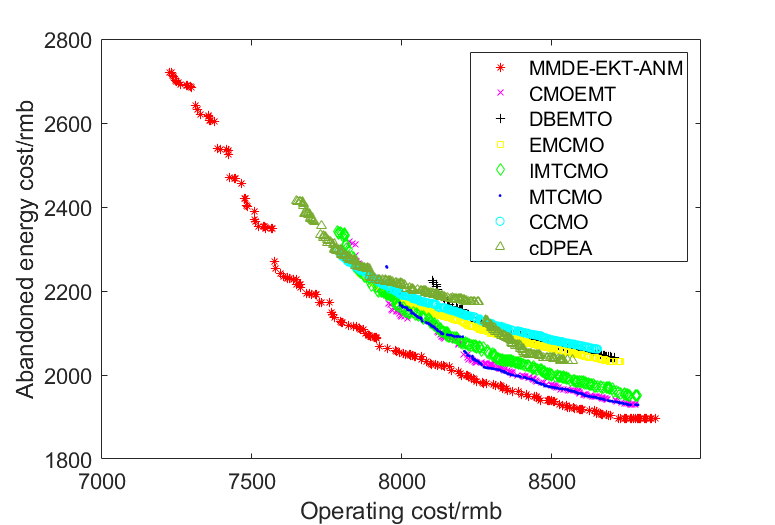}	
\caption{The PF comparison between the proposed algorithm and its competitors}
\label{ALL1}
\end{figure}

 To visually demonstrate the distribution of feasible solution sets obtained by all algorithms, we plot the PF comparison chart as shown in Figure \ref{ALL1}. The figure demonstrates that the proposed algorithm outperforms other compared algorithms in terms of diversity and convergence, which is consistent with the evaluation results of the IGD and HV indicators recorded in Table \ref{IGD} and Table \ref{HV}.

\subsection{Validation of the proposed strategy's effectiveness}\label{sec5.4}
\par
\begin{figure}[!htb]
\centering	
\includegraphics[width=8cm, height=6cm]{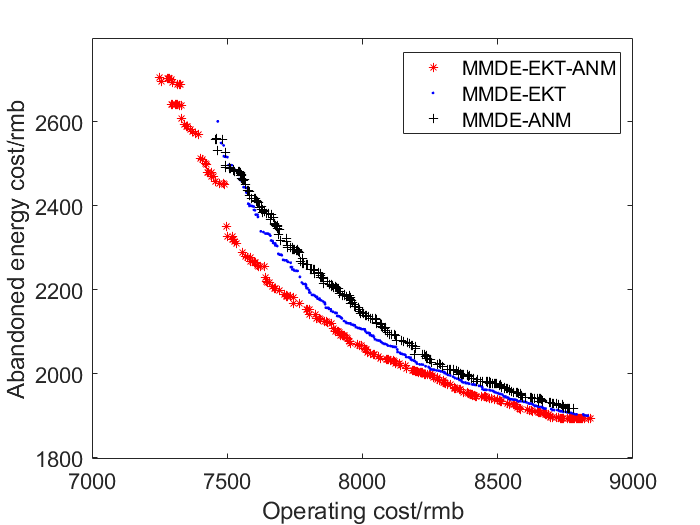}	
\caption{The PF comparison between the proposed algorithm and its variant algorithms}
\label{xiaorong1}
\end{figure}

 To validate the effectiveness of the adaptive neighborhood mutation and elite-guided knowledge transfer strategies in the MMDE-EKT-ANM algorithm, we design two variant algorithms, namely MMDE-EKT and MMDE-ANM. In the MMDE-EKT variant, the neighborhood strategy is removed, while in the MMDE-ANM variant, individual selection based on special crowding distance is replaced with random individual selection. Subsequently, the IGD and HV indicators are employed to compare the performance of the two variant algorithms against the MMDE-EKT-ANM algorithm. The comparison results are presented in Table \ref{variant}. And the visualized results are demonstrated in Figure \ref{xiaorong1}. From Table \ref{variant} it can be also concluded that MMDE-EKT-ANM achieves the most favorable outcomes in terms of the Mean, Std, Worst, and Best values of the IGD and HV indicators, indicating its superiority in terms of convergence, diversity, and stability. Figure \ref{xiaorong1} indicates that the proposed algorithm surpasses the two variant algorithms in terms of both diversity and convergence. In summary, the effectiveness of the proposed strategy can be clearly proved by the experimental results of two performance indicators and PF distribution.

\par Accordingly, the reason that MMDE-EKT-ANM outperforms other compared evolutionary algorithms lies in the following three points. Firstly, the proposed multi-task multi-constraint evolutionary dispatch algorithm with the multi-task construction strategy based on constraint relationship analysis can effectively reduce the difficulty of constraint handling. Secondly, the selection of transfer individuals in the elite-guided knowledge transfer strategy takes into account the diversity of objective space and decision space, so as to obtain a pareto front with good diversity and convergence. Thirdly, the designed adaptive neighborhood mutation mechanism improves the performance of DE, allowing DE to escape locally feasible regions and obtain a globally optimal feasible solution set. Therefore, these factors contribute to the outstanding performance of MMDE-EKT-ANM, setting it apart from other algorithms and making it more effective for solving dispatch optimization challenges in coal mine integrated energy systems.

\begin{table}[!htb]
\footnotesize
  \caption{The IGD and HV values of the proposed algorithm and its variants }\label{variant}
  \begin{tabular}{llllllll}
	\hline
Metric &Algorithm &MMDE-EKT-ANM&MMDE-EKT&MMDE-ANM\\[1pt]

\hline
    \multirow{4}{*}{IGD}
    &Mean &\textbf{19.1998} &\underline{48.1238} &68.2885\\
    & Std  & \textbf{6.5863} &8.6764  & \underline{8.5805} \\
    & Worst  & \textbf{30.4183} & \underline{62.3332} & 82.7148  \\
    & Best  &\textbf{6.6873} & \underline{29.3833} & 50.9137  \\
\hline
    \multirow{4}{*}{HV}
    &Mean &\textbf{0.9653} 	&\underline{0.9178 }    &0.8847\\
    & Std &\underline{0.0117 }	&0.0133 	&\textbf{0.0112}\\
    & Worst &\textbf{0.9893} 	&\underline{0.9464 }	&0.9085\\
    & Worst &\textbf{0.9468} 	&\underline{0.8931 }	&0.8686\\

    \bottomrule
  \end{tabular}
\end{table}

\subsection{Analysis of dispatch results}\label{sec5.5}
\par
\begin{figure}[!htb]
\centering	
\includegraphics[width=8cm, height=4.5cm]{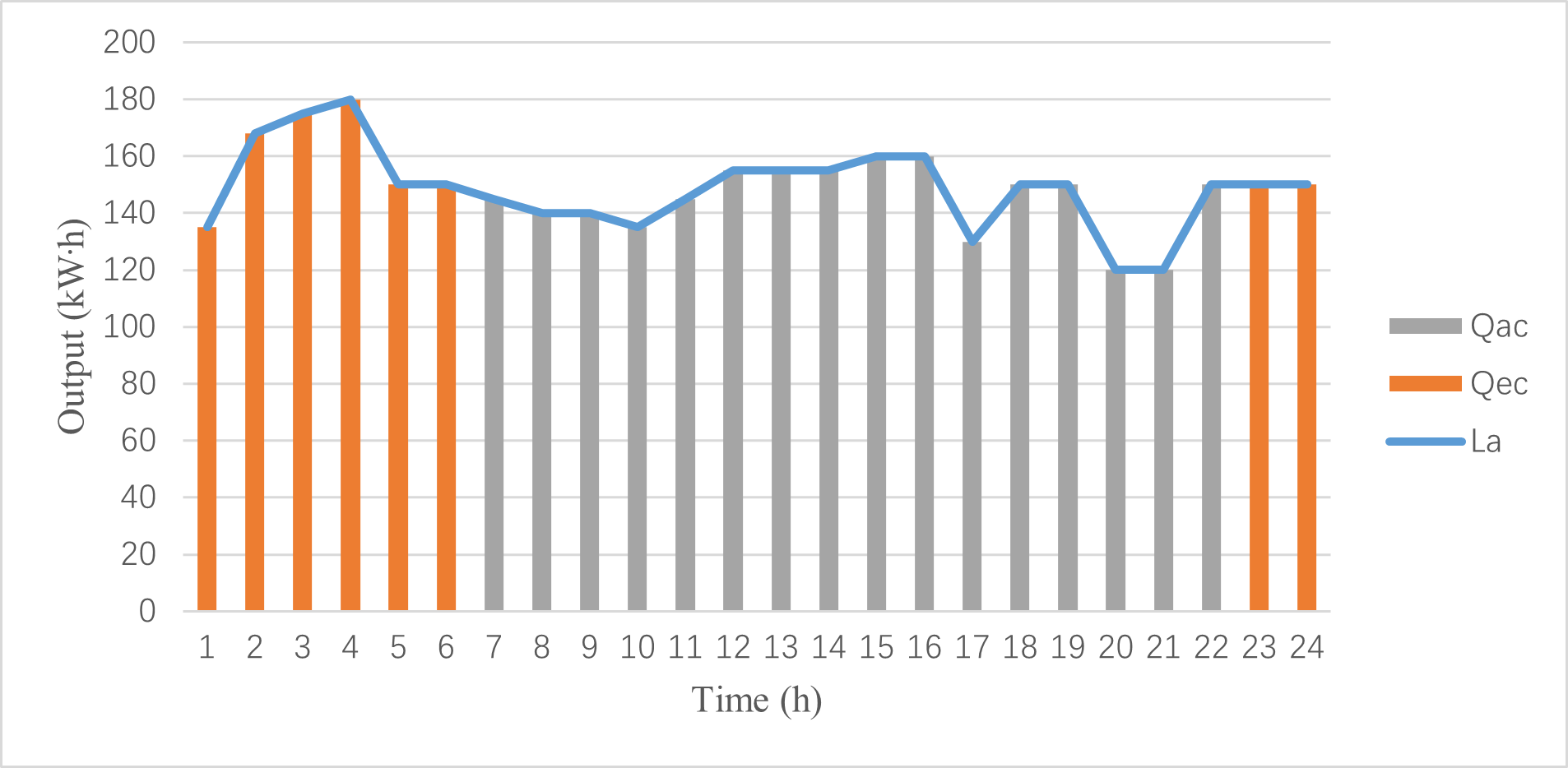}	
\caption{Cooling dispatch result}
\label{C-dispatch}
\end{figure}

\begin{figure}[!htb]
\centering	
\includegraphics[width=8cm, height=4.5cm]{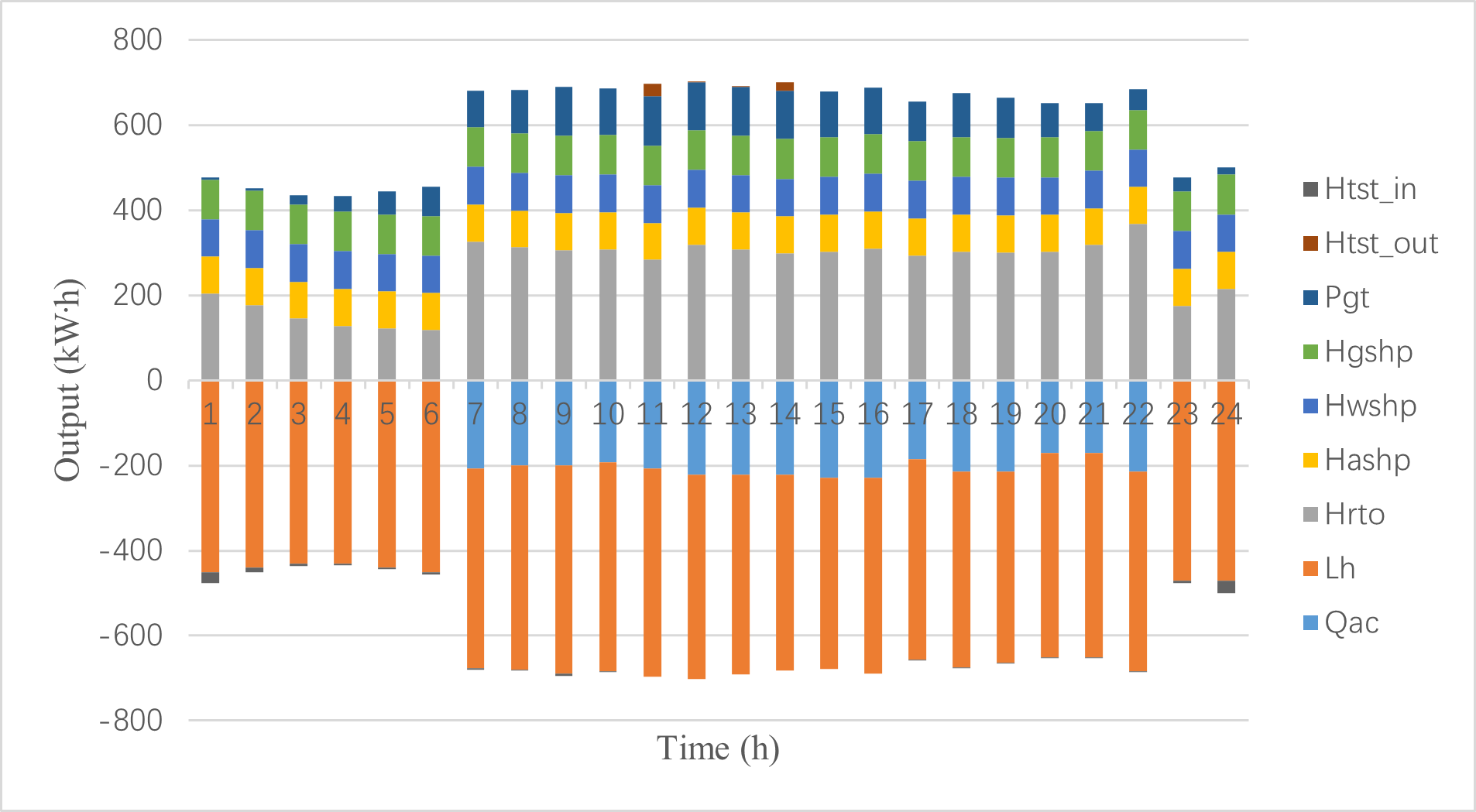}	
\caption{Thermal dispatch result}
\label{H-dispatch}
\end{figure}

\begin{figure}[!htb]
\centering	
\includegraphics[width=8cm, height=4.5cm]{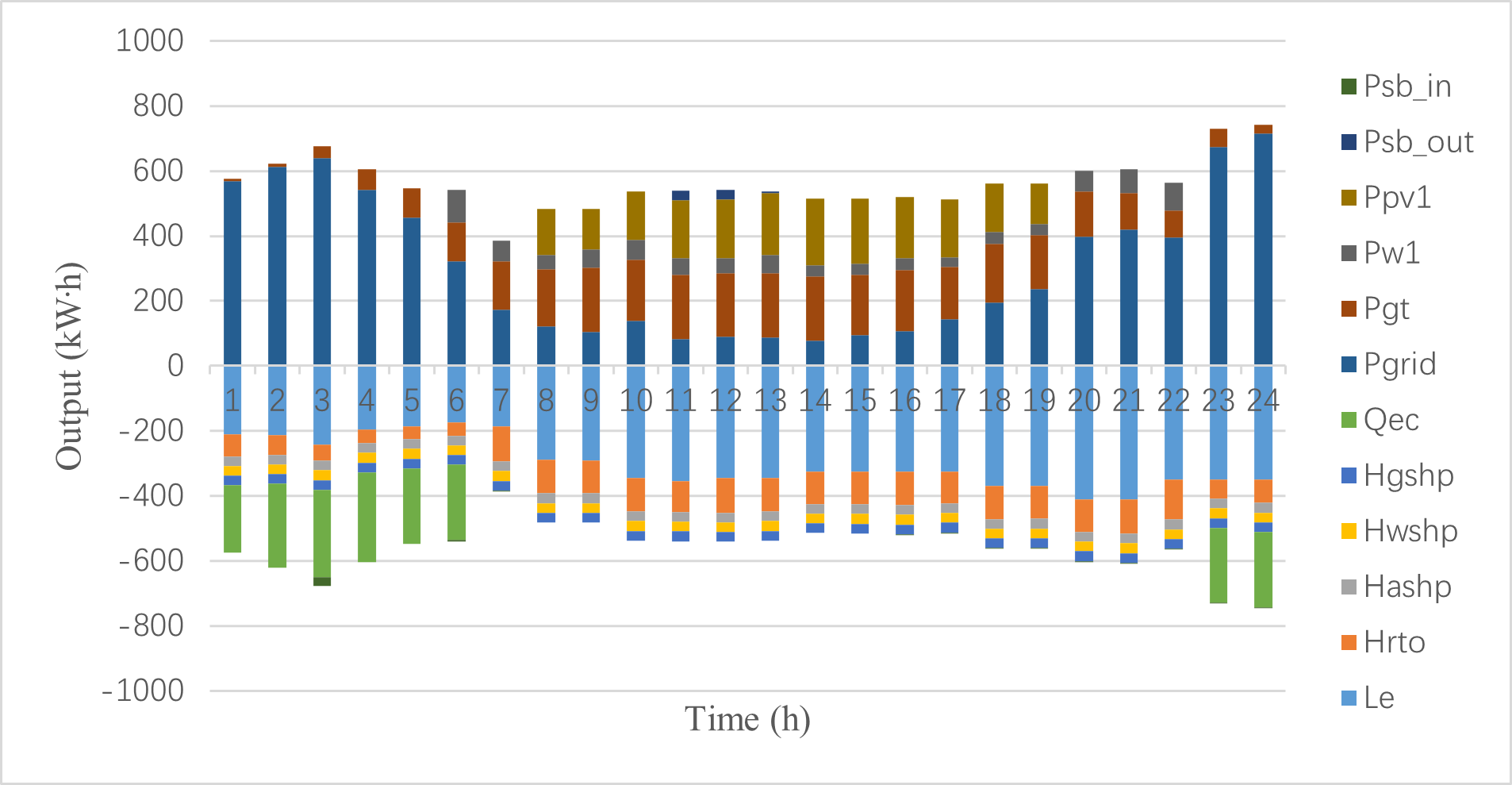}	
\caption{Electrical dispatch result}
\label{E-dispatch}
\end{figure}

 In this section, the left endpoint of PF obtained by MMDE-EKT-ANM is selected as a representative dispatch scenario, and an energy analysis is conducted to demonstrate the feasibility of the obtained results in practical applications. The dispatch results for cooling, thermal, and electrical are illustrated in Figures \ref{C-dispatch}-\ref{E-dispatch}. (1) Figure \ref{C-dispatch} reveals that during the periods of 1-6h and 23-24h, the cooling load is exclusively supplied by electrical chiller, benefiting from lower electricity prices. However, in the periods of 7-22h, the cooling load is entirely met by absorption chiller due to higher electricity prices. (2) From Figure \ref{H-dispatch}, throughout the dispatch period, the output power of the ventilation air methane oxidation devices is higher than that of the other three associated energy devices. This is primarily due to the fact that considering environmental pollution, the penalty coefficient of abandoned ventilation air methane is higher than that of the other three associated energy sources. Consequently, the ventilation air methane is prioritized for consumption to reduce the abandoned energy cost. During the 7-22h, the absorption chiller consumes thermal power, increasing the thermal demand. To maintain thermal power balance, the increased thermal demand leads to an increase in the output power of the associated energy devices and the gas turbine. (3) Figure \ref{E-dispatch} shows that during the 1-5h and 23-24h, which are characterized by low electricity demand, the electricity is primarily supplied by the power grid and the gas turbine. During the 6-7h and 20-22h, as electricity prices increase, the output power of the power grid decreases, and wind power begins to participate in the system dispatch. During the 8-19h, because the operating costs of the gas turbine and photovoltaic are relatively low, they are given priority for electricity generation, with the power grid and wind power serving as supplementary sources. In summary, the energy input/output of each device in this scenario satisfies the constraints of supply and demand balance and accords with the actual operation demand, thus verifying the feasibility of the results obtained by MMDE-EKT-ANM.

\section{Conclusion}\label{sec6}
\par A domain knowledge-driven multi-task multi-constraint evolutionary algorithm is presented to effectively solve the dispatch of coal mine integrated energy system with great number of optimized variables and strongly coupled constraints. From the viewpoint of domain-adaptive task construction, three modes are conveyed by deeply analyzing the constraint relationships. Under the constructed multi-task optimization framework, an enhanced differential evolution algorithm articulated with elite-guided knowledge transfer strategy and adaptive neighborhood mutation technique is developed. The algorithm is applied to a practical coal mine integrated energy system, and its performance in obtaining results with outstanding convergence, diversity, stability and feasibility is sufficiently demonstrated by comparing with CPLEX and seven state-of-the-art evolutionary multi-objective algorithms.

Compared to the CPLEX solver, our algorithm is time consuming. Besides, the proposed algorithm may encounter scalability challenges when solving operational optimization problems in uncertain scenarios. These problems will be further studied in the future.

\section*{Acknowledgements}
This work was supported in part by the National Natural Science Foundation of China (No. 62133015) and  the  National Key R\&D Program of China (No. 2021YFE0199000).

\bibliographystyle{unsrt}
\bibliography{mybibfile}

\begin{thebibliography}{10}

\bibitem{wu2016integrated}
Jianzhnog Wu, Jinyue Yan, Hongjie Jia, Nikos Hatziargyriou, Ned Djilali, and
  Hongbin Sun.
\newblock Integrated energy systems.
\newblock {\em Applied Energy}, 167:155--157, 2016.

\bibitem{boza2021artificial}
Pal Boza and Theodoros Evgeniou.
\newblock Artificial intelligence to support the integration of variable
  renewable energy sources to the power system.
\newblock {\em Applied Energy}, 290:116754, 2021.

\bibitem{hu2022enhanced}
Hejuan Hu, Xiaoyan Sun, Bo~Zeng, Dunwei Gong, and Yong Zhang.
\newblock Enhanced evolutionary multi-objective optimization-based dispatch of
  coal mine integrated energy system with flexible load.
\newblock {\em Applied Energy}, 307:118130, 2022.

\bibitem{zhou2019operation}
Cheng Zhou, Jianyong Zheng, Sai Liu, Yu~Liu, Fei Mei, Yi~Pan, Tian Shi, and
  Jianzhang Wu.
\newblock Operation optimization of multi-district integrated energy system
  considering flexible demand response of electric and thermal loads.
\newblock {\em Energies}, 12(20):3831, 2019.

\bibitem{wang2019operation}
Yongli Wang, Yudong Wang, Yujing Huang, Jiale Yang, Yuze Ma, Haiyang Yu, Ming
  Zeng, Fuwei Zhang, and Yanfu Zhang.
\newblock Operation optimization of regional integrated energy system based on
  the modeling of electricity-thermal-natural gas network.
\newblock {\em Applied Energy}, 251:113410, 2019.

\bibitem{zhang2021day}
Zhenwei Zhang, Chengfu Wang, Huacan Lv, Fengquan Liu, Hongzhang Sheng, and Ming
  Yang.
\newblock Day-ahead optimal dispatch for integrated energy system considering
  power-to-gas and dynamic pipeline networks.
\newblock {\em IEEE Transactions on Industry Applications}, 57(4):3317--3328,
  2021.

\bibitem{song2021economic}
Xiaoling Song, Yudong Wang, Zhe Zhang, Charles Shen, and Feniosky
  Pe{\~n}a-Mora.
\newblock Economic-environmental equilibrium-based bi-level dispatch strategy
  towards integrated electricity and natural gas systems.
\newblock {\em Applied Energy}, 281:116142, 2021.

\bibitem{zhang2021multi}
Dongdong Zhang, Hongyu Zhu, Hongcai Zhang, Hui~Hwang Goh, Hui Liu, and Thomas
  Wu.
\newblock Multi-objective optimization for smart integrated energy system
  considering demand responses and dynamic prices.
\newblock {\em IEEE Transactions on Smart Grid}, 13(2):1100--1112, 2021.

\bibitem{wang2018optimal}
Yongli Wang, Yudong Wang, Yujing Huang, Haiyang Yu, Ruiting Du, Fuli Zhang,
  Fuwei Zhang, and Jinrong Zhu.
\newblock Optimal scheduling of the regional integrated energy system
  considering economy and environment.
\newblock {\em IEEE Transactions on Sustainable Energy}, 10(4):1939--1949,
  2018.

\bibitem{wu2023multi}
Xiaonan Wu, Borui Liao, Yaogang Su, and Shuang Li.
\newblock Multi-objective and multi-algorithm operation optimization of
  integrated energy system considering ground source energy and solar energy.
\newblock {\em International Journal of Electrical Power \& Energy Systems},
  144:108529, 2023.

\bibitem{li2018multi}
Guozheng Li, Rui Wang, Tao Zhang, and Mengjun Ming.
\newblock Multi-objective optimal design of renewable energy integrated cchp
  system using picea-g.
\newblock {\em Energies}, 11(4):743, 2018.

\bibitem{wu2021multitasking}
Ting Wu, Siqi Bu, Xiang Wei, Guibin Wang, and Bin Zhou.
\newblock Multitasking multi-objective operation optimization of integrated
  energy system considering biogas-solar-wind renewables.
\newblock {\em Energy conversion and management}, 229:113736, 2021.

\bibitem{dong2023hybrid}
Yingchao Dong, Hongli Zhang, Ping Ma, Cong Wang, and Xiaojun Zhou.
\newblock A hybrid robust-interval optimization approach for integrated energy
  systems planning under uncertainties.
\newblock {\em Energy}, 274:127267, 2023.

\bibitem{wu2021multi}
Jun Wu, Baolin Li, Jun Chen, Yurong Ding, Qinghui Lou, Xiangyu Xing, Peiying
  Gan, Hui Zhou, and Junfeng Chen.
\newblock Multi-objective optimal scheduling of offshore micro integrated
  energy system considering natural gas emission.
\newblock {\em International Journal of Electrical Power \& Energy Systems},
  125:106535, 2021.

\bibitem{wang2020economic}
Yongli Wang, Yuze Ma, Fuhao Song, Yang Ma, Chengyuan Qi, Feifei Huang, Juntai
  Xing, and Fuwei Zhang.
\newblock Economic and efficient multi-objective operation optimization of
  integrated energy system considering electro-thermal demand response.
\newblock {\em Energy}, 205:118022, 2020.

\bibitem{wang2022unified}
Yan Wang, Hejuan Hu, Xiaoyan Sun, Yong Zhang, and Dunwei Gong.
\newblock Unified operation optimization model of integrated coal mine energy
  systems and its solutions based on autonomous intelligence.
\newblock {\em Applied Energy}, 328:120106, 2022.

\bibitem{dai2023constraint}
Canyun Dai, Xiaoyan Sun, Hejuan Hu, Yong Zhang, and Dunwei Gong.
\newblock A constraint adaptive multi-tasking differential evolution algorithm:
  Designed for dispatch of integrated energy system in coal mine.
\newblock {\em Tsinghua Science and Technology}, 29(2):368--385, 2023.

\bibitem{gupta2015multifactorial}
Abhishek Gupta, Yew-Soon Ong, and Liang Feng.
\newblock Multifactorial evolution: toward evolutionary multitasking.
\newblock {\em IEEE Transactions on Evolutionary Computation}, 20(3):343--357,
  2015.

\bibitem{woldesenbet2009constraint}
Yonas~Gebre Woldesenbet, Gary~G Yen, and Biruk~G Tessema.
\newblock Constraint handling in multiobjective evolutionary optimization.
\newblock {\em IEEE Transactions on Evolutionary Computation}, 13(3):514--525,
  2009.

\bibitem{deb2000efficient}
Kalyanmoy Deb.
\newblock An efficient constraint handling method for genetic algorithms.
\newblock {\em Computer methods in applied mechanics and engineering},
  186(2-4):311--338, 2000.

\bibitem{takahama2010efficient}
Tetsuyuki Takahama and Setsuko Sakai.
\newblock Efficient constrained optimization by the $\varepsilon$ constrained
  adaptive differential evolution.
\newblock {\em IEEE congress on evolutionary computation}, 2010.

\bibitem{tasgetiren2010ensemble}
M~Fatih Tasgetiren, P~Nagaratnam Suganthan, Quan-Ke Pan, Rammohan Mallipeddi,
  and Sedat Sarman.
\newblock An ensemble of differential evolution algorithms for constrained
  function optimization.
\newblock {\em IEEE congress on evolutionary computation}, 2010.

\bibitem{qiao2022dynamic}
Kangjia Qiao, Kunjie Yu, Boyang Qu, Jing Liang, Hui Song, Caitong Yue, Hongyu
  Lin, and Kay~Chen Tan.
\newblock Dynamic auxiliary task-based evolutionary multitasking for
  constrained multi-objective optimization.
\newblock {\em IEEE Transactions on Evolutionary Computation}, 2022.

\bibitem{yue2017multiobjective}
Caitong Yue, Boyang Qu, and Jing Liang.
\newblock A multiobjective particle swarm optimizer using ring topology for
  solving multimodal multiobjective problems.
\newblock {\em IEEE Transactions on Evolutionary Computation}, 22(5):805--817,
  2017.

\bibitem{storn1997differential}
Rainer Storn and Kenneth Price.
\newblock Differential evolution--a simple and efficient heuristic for global
  optimization over continuous spaces.
\newblock {\em Journal of global optimization}, 11:341--359, 1997.

\bibitem{sheng2020differential}
Weiguo Sheng, Xi~Wang, Zidong Wang, Qi~Li, Yujun Zheng, and Shengyong Chen.
\newblock A differential evolution algorithm with adaptive niching and k-means
  operation for data clustering.
\newblock {\em IEEE Transactions on Cybernetics}, 52(7):6181--6195, 2020.

\bibitem{wang2022surrogate}
Weizhong Wang, Hai-Lin Liu, and Kay~Chen Tan.
\newblock A surrogate-assisted differential evolution algorithm for
  high-dimensional expensive optimization problems.
\newblock {\em IEEE Transactions on Cybernetics}, 53(4):2685--2697, 2022.

\bibitem{xiao2023locally}
Leyi Xiao, Chaodong Fan, Zhaoyang Ai, and Jie Lin.
\newblock Locally informed gravitational search algorithm with hierarchical
  topological structure.
\newblock {\em Engineering Applications of Artificial Intelligence},
  123:106236, 2023.

\bibitem{zhou2023clustering}
Ting Zhou, Zhongbo Hu, Qinghua Su, and Wentao Xiong.
\newblock A clustering differential evolution algorithm with neighborhood-based
  dual mutation operator for multimodal multiobjective optimization.
\newblock {\em Expert Systems with Applications}, 216:119438, 2023.

\bibitem{tian2017platemo}
Ye~Tian, Ran Cheng, Xingyi Zhang, and Yaochu Jin.
\newblock Platemo: A matlab platform for evolutionary multi-objective
  optimization [educational forum].
\newblock {\em IEEE Computational Intelligence Magazine}, 12(4):73--87, 2017.

\bibitem{tian2020coevolutionary}
Ye~Tian, Tao Zhang, Jianhua Xiao, Xingyi Zhang, and Yaochu Jin.
\newblock A coevolutionary framework for constrained multiobjective
  optimization problems.
\newblock {\em IEEE Transactions on Evolutionary Computation}, 25(1):102--116,
  2020.

\bibitem{ming2021dual}
Mengjun Ming, Anupam Trivedi, Rui Wang, Dipti Srinivasan, and Tao Zhang.
\newblock A dual-population-based evolutionary algorithm for constrained
  multiobjective optimization.
\newblock {\em IEEE Transactions on Evolutionary Computation}, 25(4):739--753,
  2021.

\bibitem{qiao2022evolutionary}
Kangjia Qiao, Kunjie Yu, Boyang Qu, Jing Liang, Hui Song, and Caitong Yue.
\newblock An evolutionary multitasking optimization framework for constrained
  multiobjective optimization problems.
\newblock {\em IEEE Transactions on Evolutionary Computation}, 26(2):263--277,
  2022.

\bibitem{ming2022constrained}
Fei Ming, Wenyin Gong, Ling Wang, and Liang Gao.
\newblock Constrained multi-objective optimization via multitasking and
  knowledge transfer.
\newblock {\em IEEE Transactions on Evolutionary Computation}, 2022.

\bibitem{qiao2023self}
Kangjia Qiao, Jing Liang, Kunjie Yu, Minghui Wang, Boyang Qu, Caitong Yue, and
  Yinan Guo.
\newblock A self-adaptive evolutionary multi-task based constrained
  multi-objective evolutionary algorithm.
\newblock {\em IEEE Transactions on Emerging Topics in Computational
  Intelligence}, 2023.

\bibitem{qiao2023evolutionary}
Kangjia Qiao, Jing Liang, Kunjie Yu, Caitong Yue, Hongyu Lin, Dezheng Zhang,
  and Boyang Qu.
\newblock Evolutionary constrained multiobjective optimization: Scalable
  high-dimensional constraint benchmarks and algorithm.
\newblock {\em IEEE Transactions on Evolutionary Computation}, 2023.

\bibitem{bosman2003balance}
Peter~AN Bosman and Dirk Thierens.
\newblock The balance between proximity and diversity in multiobjective
  evolutionary algorithms.
\newblock {\em IEEE transactions on evolutionary computation}, 7(2):174--188,
  2003.

\bibitem{liu2020fuzzy}
Songbai Liu, Qiuzhen Lin, Kay~Chen Tan, Maoguo Gong, and Carlos A~Coello
  Coello.
\newblock A fuzzy decomposition-based multi/many-objective evolutionary
  algorithm.
\newblock {\em IEEE transactions on cybernetics}, 52(5):3495--3509, 2020.

\end{thebibliography}

\end{document}